%% file: eccv2020submission.tex
\documentclass[runningheads]{llncs}
\usepackage{graphicx}
\usepackage{comment}
\usepackage{amsmath,amssymb} 
\usepackage{color}

\usepackage{times}
\usepackage{epsfig}
\usepackage{graphicx}
\usepackage{amsmath}
\usepackage{amssymb}
\usepackage{siunitx}
\usepackage{mycommands}
\usepackage{color}
\usepackage{pifont}
\newcommand{\cmark}{\ding{51}}%
\newcommand{\xmark}{\ding{55}}
\usepackage{tabularx}
\usepackage{xpatch}

\usepackage{floatrow}

\usepackage[width=122mm,left=12mm,paperwidth=146mm,height=193mm,top=12mm,paperheight=217mm]{geometry}

\usepackage[pagebackref=true,breaklinks=true,letterpaper=true,colorlinks,bookmarks=false]{hyperref}

\begin{document}
\pagestyle{headings}
\mainmatter

\input{sections/title.tex}

\input{sections/abstract.tex}


\input{sections/intro.tex}

\input{sections/rw.tex}
\input{sections/method.tex}

\input{sections/experiments.tex}

\input{sections/conclusion.tex}
\input{sections/training_cost.tex}
\input{sections/parameters.tex}
\input{sections/gldv2_trained_results.tex}
\input{sections/memory.tex}
\input{sections/viz_results.tex}
\input{sections/viz_dataset_examples.tex}
\input{sections/viz_optimization.tex}

%
%
\bibliographystyle{splncs04}
\bibliography{literature/delf}

\end{document}

%% file: sections/title.tex

\title{Unifying Deep Local and Global Features for \\Image Search} 


\titlerunning{Unifying Deep Local and Global Features for Image Search}
%
\author{Bingyi Cao\thanks{Both authors contributed equally to this work.}\quad Andr\'{e} Araujo\protect\footnotemark[1] \quad Jack Sim}
\authorrunning{B. Cao et al.}
%
\institute{Google Research, USA\\
\email{\{bingyi,andrearaujo,jacksim\}@google.com}}
\maketitle

%% file: sections/abstract.tex

\begin{abstract}

Image retrieval is the problem of searching an image database for items that are similar to a query image.
To address this task, two main types of image representations have been studied: global and local image features.
In this work, our key contribution is to unify global and local features into a single deep model, enabling accurate retrieval with efficient feature extraction.
We refer to the new model as DELG, standing for DEep Local and Global features.
We leverage lessons from recent feature learning work and propose a model that combines generalized mean pooling for global features and attentive selection for local features.
The entire network can be learned end-to-end by carefully balancing the gradient flow between two heads -- requiring only image-level labels.
We also introduce an autoencoder-based dimensionality reduction technique for local features, which is integrated into the model, improving training efficiency and matching performance.
Comprehensive experiments show that our model achieves state-of-the-art image retrieval on the Revisited Oxford and Paris datasets, and state-of-the-art single-model instance-level recognition on the Google Landmarks dataset v2.
Code and models are available at \url{https://github.com/tensorflow/models/tree/master/research/delf}.

\keywords{deep features, image retrieval, unified model}

\end{abstract}

%% file: sections/intro.tex

\section{Introduction}

Large-scale image retrieval is a long-standing problem in computer vision, which saw promising results \cite{Nister,Philbin07,Philbin2008,jegou2012aggregating} even before deep learning revolutionized the field.
Central to this problem are the representations used to describe images and their similarities.

\begin{figure}[t]
\begin{center}
   \includegraphics[width=1.0\linewidth]{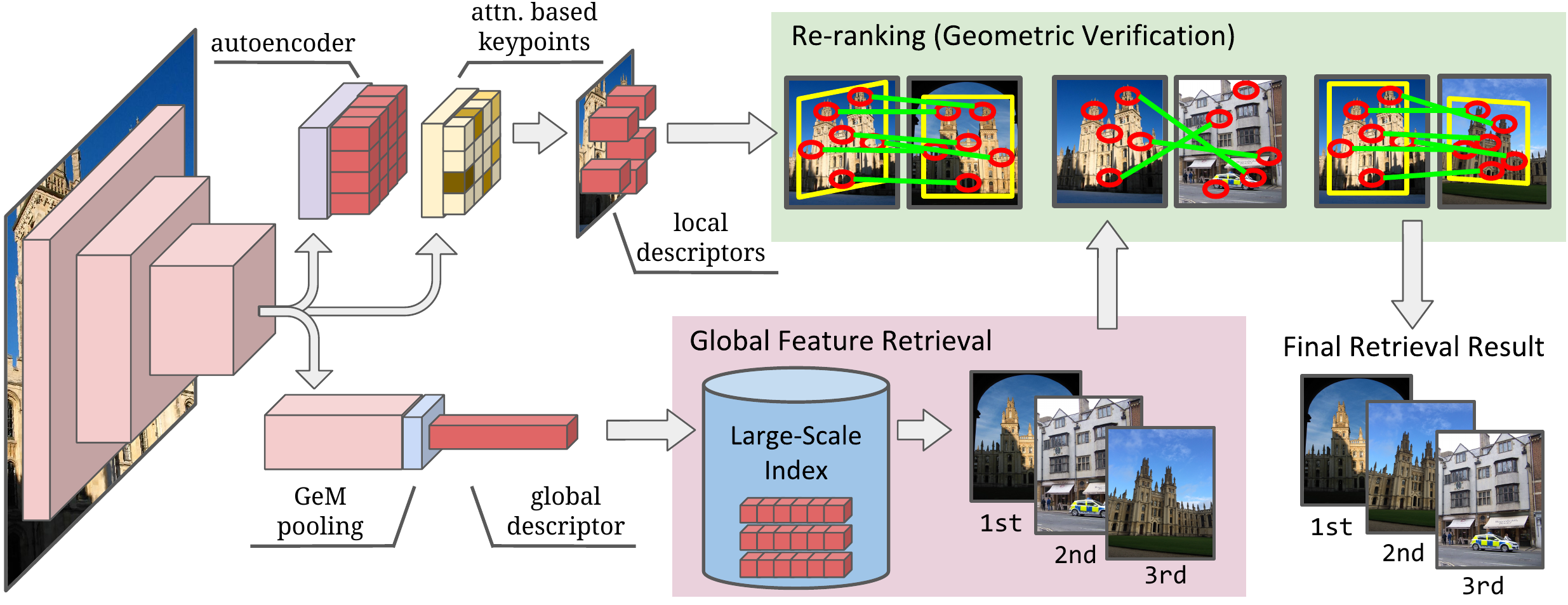}
\end{center}
   \caption{
Our proposed \textbf{DELG (DEep Local and Global features)} model (left) jointly extracts deep local and global features. Global features can be used in the first stage of a retrieval system, to efficiently select the most similar images (bottom). Local features can then be employed to re-rank top results (top-right), increasing precision of the system. The unified model leverages hierarchical representations induced by convolutional neural networks to learn local and global features, combined with recent advances in global pooling and attentive local feature detection.
   }
\label{fig:key_fig}
\end{figure}

Two types of image representations are necessary for high image retrieval performance: global and local features.
A global feature \cite{jegou2012aggregating,arandjelovic2016netvlad,gordo2017end,radenovic2018fine,revaud2019learning}, also commonly referred to as ``global descriptor'' or ``embedding'', summarizes the contents of an image, often leading to a compact representation; information about spatial arrangement of visual elements is lost.
Local features \cite{Lowe2004,bay2008speeded,yi2016lift,noh2017large,mishkin2018repeatability}, on the other hand, comprise descriptors and geometry information about specific image regions; they are especially useful to match images depicting rigid objects.
Generally speaking, global features are better at recall, while local features are better at precision.
Global features can learn similarity across very different poses where local features would not be able to find correspondences; in contrast,
the score provided by local feature-based geometric verification usually reflects image similarity well, being more reliable than global feature distance.
A common retrieval system setup is to first search by global features, then re-rank the top database images using local feature matching -- to get the best of both worlds.
Such a hybrid approach gained popularity in visual localization \cite{taira2018inloc,sarlin2019from} and instance-level recognition problems \cite{ozaki2019large,weyand2020google}.

Today, most systems that rely on both these types of features need to separately extract each of them, using different models.
This is undesirable since it may lead to high memory usage and increased latency, \eg{}, if both models require specialized and limited hardware such as GPUs.
Besides, in many cases similar types of computation are performed for both, resulting in redundant processing and unnecessary complexity.

\paragraph{Contributions.}
(1) Our first contribution is a unified model to represent both local and global features, using a convolutional neural network (CNN), referred to as DELG (DEep Local and Global features) -- illustrated in \figref{fig:key_fig}.
This allows for efficient inference by extracting an image's global feature, detected keypoints and local descriptors within a single model.
Our model is enabled by leveraging hierarchical image representations that arise in CNNs \cite{zeiler2014visualizing}, which we couple to generalized mean pooling \cite{radenovic2018fine} and attentive local feature detection \cite{noh2017large}.
(2) Second, we adopt a convolutional autoencoder module that can successfully learn low-dimensional local descriptors.
This can be readily integrated into the unified model, and avoids the need of post-processing learning steps, such as PCA, that are commonly used.
(3) Finally, we design a procedure that enables end-to-end training of the proposed model using only image-level supervision.
This requires carefully controlling the gradient flow between the global and local network heads during backpropagation, to avoid disrupting the desired representations.
Through systematic experiments, we show that our joint model achieves state-of-the-art performance on the Revisited Oxford, Revisited Paris and Google Landmarks v2 datasets.


%
%

%% file: sections/rw.tex

\section{Related Work}

We review relevant work in local and global features, focusing mainly on approaches related to image retrieval.

\textbf{Local features.}
Hand-crafted techniques such as SIFT \cite{Lowe2004} and SURF \cite{bay2008speeded} have been widely used for retrieval problems.
Early systems \cite{Mikolajczyk2002affine,Lowe2004,obdrzalek2005sublinear} worked by searching for query local descriptors against a large database of local descriptors, followed by geometrically verifying database images with sufficient number of correspondences.
Bag-of-Words \cite{Sivic2003} and related methods \cite{Philbin07,Philbin2008,Jegou2008} followed, by relying on visual words obtained via local descriptor clustering, coupled to TF-IDF scoring.
The key advantage of local features over global ones for retrieval is the ability to perform spatial matching, often employing RANSAC \cite{Fischler1981}.
This has been widely used \cite{Philbin07,Philbin2008,avrithis2014hough}, as it produces reliable and interpretable scores.
Recently, several deep learning-based local features have been proposed \cite{yi2016lift,noh2017large,mishchuk2017working,ono2018lfnet,mishkin2018repeatability,revaud2019r2d2,Dusmanu2019CVPR,barroso2019key,luo2019contextdesc}.
The one most related to our work is DELF \cite{noh2017large}; our proposed unified model incorporates DELF's attention module, but with a much simpler training pipeline, besides also enabling global feature extraction.

\textbf{Global features}
excel at delivering high image retrieval performance with compact representations.
Before deep learning was popular in computer vision, they were developed mainly by aggregating hand-crafted local descriptors \cite{jegou2010aggregating,jegou2012aggregating,jegou2014triangulation,tolias2015image}.
Today, most high-performing global features are based on deep convolutional neural networks \cite{babenko2014neural,babenko2015iccv,tolias2015particular,arandjelovic2016netvlad,gordo2017end,radenovic2018fine,revaud2019learning}, which are trained with ranking-based \cite{chopra2005learning,schroff2015facenet,he2018local} or classification losses \cite{wang2018cosface,deng2019arcface}.
Our work leverages recent learned lessons in global feature design, by adopting GeM pooling \cite{radenovic2018fine} and ArcFace loss \cite{deng2019arcface}.
This leads to improved global feature retrieval performance compared to previous techniques, which is further boosted by geometric re-ranking with local features obtained from the same model.

\textbf{Joint local and global CNN features.}
Previous work considered neural networks for joint extraction of global and local features.
For indoor localization, Taira \etal{} \cite{taira2018inloc} used NetVLAD \cite{arandjelovic2016netvlad} to extract global features for candidate pose retrieval, followed by dense local feature matching using feature maps from the same network.
Simeoni \etal{}'s DSM \cite{simeoni2019local} detected keypoints in activation maps from global feature models using MSER \cite{matas2004robust}; activation channels are interpreted as visual words, in order to propose correspondences between a pair of images.
Our work differs substantially from \cite{taira2018inloc,simeoni2019local}, since they only post-process pre-trained global feature models to produce local features, while we jointly train local and global.
Sarlin \etal{} \cite{sarlin2019from} distill pre-trained local \cite{detone18superpoint} and global \cite{arandjelovic2016netvlad} features into a single model, targeting localization applications.
In contrast, our model is trained end-to-end for image retrieval, and is not limited to mimicking separate pre-trained local and global models. 
To the best of our knowledge, ours is the first work to learn a non-distilled model producing both local and global features.

\textbf{Dimensionality reduction for image retrieval.}
PCA and whitening are widely used for dimensionality reduction of local and global features in image retrieval \cite{babenko2015iccv,tolias2015particular,noh2017large,revaud2019learning}.
As discussed in \cite{jegou2012negative}, whitening downweights co-occurences of local features, which is generally beneficial for retrieval applications.
Mukundan \etal{} \cite{mukundan2019understanding} further introduce a shrinkage parameter that controls the extent of applied whitening.
If supervision in the form of matching pairs or category labels is available, more sophisticated methods \cite{mikolajczyk2007improving,gordo2012leveraging} can be used.
More recently, Gordo \etal{} \cite{gordo2016deep} propose to replace PCA/whitening by a fully-connected layer, that is learned together with the global descriptor.

In this paper, our goal is to compose a system that can be learned end-to-end, using only image-level labels and without requiring post-processing stages that make training more complex.
Also, since we extract local features from feature maps of common CNN backbones, they tend to be very high-dimensional and infeasible for large-scale problems.
All above-mentioned approaches would either require a separate post-processing step to reduce the dimensionality of features, or supervision at the level of local patches -- making them unsuitable to our needs.
We thus introduce an autoencoder in our model, which can be jointly and efficiently learned with the rest of the network.
It requires no extra supervision as it can be trained with a reconstruction loss.


%% file: sections/method.tex

\section{DELG} \label{sec:method}

\subsection{Design considerations}


For optimal performance, image retrieval requires semantic understanding of the types of objects that a user may be interested in, such that the system can distinguish between relevant objects versus clutter/background.
Both local and global features should thus focus only on the most discriminative information within the image.
However, there are substantial differences in terms of the desired behavior for these two feature modalities, posing a considerable challenge to jointly learn them.

Global features should be similar for images depicting the same object of interest, and dissimilar otherwise.
This requires high-level, abstract representations that are invariant to viewpoint and photometric transformations.
Local features, on the other hand, need to encode representations that are grounded to specific image regions; in particular, the keypoint detector should be equivariant with respect to viewpoint, and the keypoint descriptor needs to encode localized visual information.
This is crucial to enable geometric consistency checks between query and database images, which are widely used in image retrieval systems.

Besides, our goal is to design a model that can be learned end-to-end, with local and global features, without requiring additional learning stages.
This simplifies the training pipeline, allowing faster iterations and wider applicability.
In comparison, it is common for previous feature learning work to require several learning stages: attentive deep local feature learning \cite{noh2017large} requires $3$ learning stages (fine-tuning, attention, PCA); deep global features usually require two stages, \eg{}, region proposal and Siamese training \cite{gordo2017end}, or Siamese training and supervised whitening \cite{radenovic2018fine}, or ranking loss training and PCA \cite{revaud2019learning}.

\subsection{Model}

We design our DELG model, illustrated in \figref{fig:key_fig}, to fulfill the requirements outlined above.
We propose to leverage hierarchical representations from CNNs \cite{zeiler2014visualizing} in order to represent the different types of features to be learned.
While global features can be associated with deep layers representing high-level cues, local features are more suitable to intermediate layers that encode localized information.

Given an image, we apply a convolutional neural network backbone to obtain two feature maps: $\mathcal{S} \in \mathcal{R}^{H_S \times W_S \times C_S}$ and $\mathcal{D} \in \mathcal{R}^{H_D \times W_D \times C_D}$, representing shallower and deeper activations respectively, where $H,W,C$ correspond to the height, width and number of channels in each case.
For common convolutional networks, $H_D \leq H_S$, $W_D \leq W_S$ and $C_D \geq C_S$; deeper layers have spatially smaller maps, with a larger number of channels.
Let $s_{h,w} \in \mathcal{R}^{C_S}$ and $d_{h,w} \in \mathcal{R}^{C_D}$ denote features at location $h,w$ in these maps.
For common network designs, these features are non-negative since they are obtained after the ReLU non-linearity, which is the case in our method.

In order to aggregate deep activations into a global feature, we adopt generalized mean pooling (GeM) \cite{radenovic2018fine}, which effectively weights the contributions of each feature.
Another key component of global feature learning is to whiten the aggregated representation; we integrate this into our model with a fully-connected layer $F \in \mathcal{R}^{C_F \times C_D}$, with learned bias $b_F \in \mathcal{R}^{C_F}$, similar to \cite{gordo2017end}.
These two components produce a global feature $g \in \mathcal{R}^{C_F}$ that summarizes the discriminative contents of the whole image:

\begin{equation}
g = F \times \left( \frac{1}{H_D W_D} \sum_{h,w} d_{h,w}^{\hspace{1pt}p} \right)^{1/p} + b_F
\label{eq:global_features}
\end{equation}

\noindent where $p$ denotes the generalized mean power parameter, and the exponentiation $d_{h,w}^{\hspace{1pt}p}$ is applied elementwise.

Regarding local features, it is important to select only the relevant regions for matching.
This can be achieved by adopting an attention module $M$ \cite{noh2017large}, whose goal is to predict which among the extracted local features are discriminative for the objects of interest. This is performed as $\mathcal{A} = M(\mathcal{S})$, where $M$ is a small convolutional network and $\mathcal{A} \in \mathcal{R}^{H_S \times W_S}$ denotes the attention score map associated to the features from $\mathcal{S}$.


Furthermore, since hundreds to thousands of local features are commonly used, they must be represented compactly.
To do so, we propose to integrate a small convolutional autoencoder (AE) module \cite{hinton1989connectionist}, which is responsible for learning a suitable low-dimensional representation.
The local descriptors are obtained as $\mathcal{L} = T(\mathcal{S})$, where $\mathcal{L} \in \mathcal{R}^{H_S \times W_S \times C_T}$, and $T$ is the encoding part of the autoencoder, corresponding to a $1 \times 1$ convolutional layer with $C_T$ filters.
Note that, contrary to $\mathcal{S}$, the local descriptors $\mathcal{L}$ are not restricted to be non-negative.

Each extracted local feature at position $h,w$ is thus represented with a local descriptor $l_{h,w} \in \mathcal{L}$ and its corresponding keypoint detection score $a_{h,w} \in \mathcal{A}$.
Their locations in the input image are set to corresponding receptive field centers, which can be computed using the parameters of the network \cite{araujo2019computing}.

The global and local descriptors are $L_2$-normalized into $\hat{g}$ and $\hat{l}_{h,w}$, respectively.

\subsection{Training}

\begin{figure*}[t]
\begin{center}
   \includegraphics[width=1.0\linewidth]{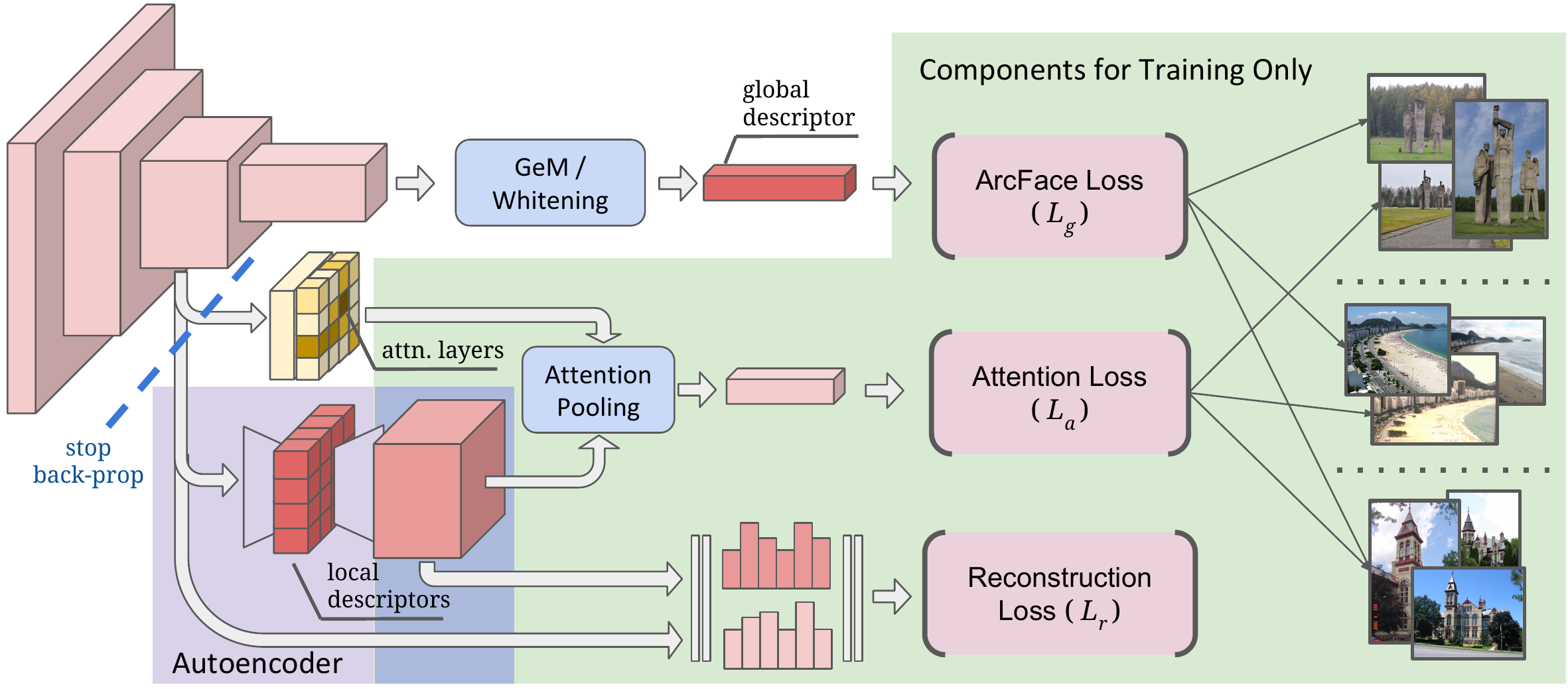}
\end{center}
   \caption{
Illustration of our \textbf{training pipeline}.
The components highlighted in green are used solely during training.
There are two classification losses: ArcFace for global feature learning ($L_g$), and softmax for attention learning ($L_a$).
In both cases, the classification objective is to distinguish different landmarks (an instance-level recognition problem).
The autoencoder (purple) is further trained with a reconstruction loss ($L_r$).
The whole model is learned end-to-end, and benefits substantially from stopping gradient back-propagation from $L_a$ and $L_r$ into the CNN backbone.
   }
\label{fig:training}
\end{figure*}

We propose to train the model using only image-level labels, as illustrated in \figref{fig:training}.
In particular, note that we do not require patch-level supervision to train local features, unlike most recent works \cite{Dusmanu2019CVPR,revaud2019r2d2,mukundan2019explicit,luo2019contextdesc}.

Besides the challenge to acquire the annotations, note that patch-level supervision could help selecting repeatable features, but not necessarily the discriminative ones; in contrast, our model discovers discriminative features by learning which can distinguish the different classes, given by image-level labels.
In this weakly-supervised local feature setting, it is very important to control the gradient flow between the global and local feature learning, which is discussed in more detail below.

\textbf{Global features.}
For global feature learning, we adopt a suitable loss function with $L_2$-normalized classifier weights $\hat{\mathcal{W}}$, followed by scaled softmax normalization and cross-entropy loss \cite{wang2017normface}; this is sometimes referred to as ``cosine classifier''.
Additionally, we adopt the ArcFace margin \cite{deng2019arcface}, which has shown excellent results for global feature learning by inducing smaller intra-class variance.
Concretely, given $\hat{g}$, we first compute the cosine similarity against $\hat{\mathcal{W}}$, adjusted by the ArcFace margin. 
The ArcFace-adjusted cosine similarity can be expressed as $\textrm{AF}(u, c)$:

\begin{equation}
\textrm{AF}(u, c) = 
\begin{cases}\!
\hspace{3pt} \textrm{cos}(\textrm{acos}(u)+m), & \text{if } c = 1 \\
\hspace{3pt} u, & \text{if } c = 0
\end{cases}
\label{eq:arcface_global}
\end{equation}

\noindent where $u$ is the cosine similarity, $m$ is the ArcFace margin and $c$ is a binary value indicating if this is the ground-truth class.
The cross-entropy loss, computed using softmax normalization can be expressed in this case as:

\begin{equation}
L_{g}(\hat{g}, y) = -\log \left( \frac{\textrm{exp}(\gamma \times \textrm{AF}(\hat{w}_k^T \hat{g}, 1))}{\sum_n \textrm{exp}(\gamma \times \textrm{AF}(\hat{w}_n^T \hat{g}, y_n))} \right)
\label{eq:softmax_global}
\end{equation}

\noindent where $\gamma$ is a learnable scalar, $\hat{w}_i$ refers to the $L_2$-normalized classifier weights for class $i$, $y$ is the one-hot label vector and $k$ is the index of the ground-truth class ($y_k=1$).

\textbf{Local features.}
To train the local features, we use two losses.
First, a mean-squared error regression loss that measures how well the autoencoder can reconstruct $\mathcal{S}$.
Denote $\mathcal{S}'=T'(\mathcal{L})$ as the reconstructed version of $\mathcal{S}$, with same dimensions, where $T'$ is a $1 \times 1$ convolutional layer with $C_S$ filters, followed by ReLU. The loss can be expressed as:

\begin{equation}
L_{r}(\mathcal{S}', \mathcal{S}) = \frac{1}{H_S W_S C_S}\sum_{h,w} \|s'_{h,w} - s_{h,w}\|^2
\label{eq:reconstruction_loss}
\end{equation}

Second, a cross-entropy classification loss that incentivizes the attention module to select discriminative local features.
This is done by first pooling the reconstructed features $\mathcal{S}'$ with attention weights $a_{h,w}$:

\begin{equation}
a'=\sum_{h,w} a_{h,w}s'_{h,w}
\label{eq:attention_pooling}
\end{equation}

\noindent Then using a standard softmax-cross-entropy loss:

\begin{equation}
L_{a}(a', k) = -\log \left( \frac{\textrm{exp}(v_k^T a' + b_k)}{\sum_n \textrm{exp}(v_n^T a' + b_n)} \right)
\label{eq:softmax_local}
\end{equation}

\noindent where $v_i,b_i$ refer to the classifier weights and biases for class $i$ and $k$ is the index of the ground-truth class; this tends to make the attention weights large for the discriminative features.
The total loss is given by $L_{g} + \lambda L_{r} + \beta L_{a}$.

\textbf{Controlling gradients.}
Naively optimizing the above-mentioned total loss experimentally leads to suboptimal results, because the reconstruction and attention loss terms significantly disturb the hierarchical feature representation which is usually obtained when training deep models.
In particular, both tend to induce the shallower features $\mathcal{S}$ to be more semantic and less localizable, which end up being sparser.
Sparser features can more easily optimize $L_{r}$, and more semantic features may help optimizing $L_{a}$; this, as a result, leads to underperforming local features.

We avoid this issue by stopping gradient back-propagation from $L_{r}$ and $L_{a}$ to the network backbone, \ie{}, to $\mathcal{S}$.
This means that the network backbone is optimized solely based on $L_{g}$, and will tend to produce the desired hierarchical feature representation.
This is further discussed in the experimental section that follows.


%% file: sections/experiments.tex

\section{Experiments} \label{sec:experiments}

\subsection{Experimental setup} \label{subsec:setup}


\textbf{Model backbone and implementation.} 
Our model is implemented using TensorFlow, leveraging the Slim model library \cite{silberman2016tensorflow}.
We use ResNet-50 (R50) and ResNet-101 (R101) \cite{he2015deep}; R50 is used for ablation experiments.
We obtain the shallower feature map $\mathcal{S}$ from the \textit{conv4} output, and the deeper feature map $\mathcal{D}$ from the \textit{conv5} output.
Note that the Slim implementation moves the \textit{conv5} stride into the last unit from \textit{conv4}, which we also adopt -- helping reduce the spatial resolution of $\mathcal{S}$.
The number of channels in $\mathcal{D}$ is $C_D=2048$; GeM pooling \cite{radenovic2018fine} is applied with parameter $p=3$, which is not learned.
The whitening fully-connected layer, applied after pooling, produces a global feature with dimensionality $C_F=2048$.
The number of channels in $\mathcal{S}$ is $C_S=1024$; the autoencoder module learns a reduced dimensionality for this feature map with $C_T=128$.
The attention network $M$ follows the setup from \cite{noh2017large}, with $2$ convolutional layers, without stride, using kernel sizes of $1$; as activation functions, the first layer uses ReLU and the second uses Softplus \cite{dugas2001incorporating}.

\textbf{Training details.}
We use the training set of the Google Landmarks dataset (GLD) \cite{noh2017large}, containing $1.2$M images from $15$k landmarks, and divide it into two subsets `train'/`val' with $80\%$/$20\%$ split.
The `train' split is used for the actual learning, and the `val' split is used for validating the learned classifier as training progresses.
Models are initialized from pre-trained ImageNet weights.
The images first undergo augmentation, by randomly cropping / distorting the aspect ratio; then, they are resized to $512\times 512$ resolution.
We use a batch size of $16$, and train using $21$ Tesla P100 GPUs asynchronously, for $1.5$M steps (corresponding to approximately $25$ epochs of the `train' split).
The model is optimized using SGD with momentum of $0.9$, and a linearly decaying learning rate that reaches zero once the desired number of steps is reached.
We experiment with initial learning rates within $[3\times 10^{-4}, 10^{-2}]$ and report results for the best performing one.
We set the ArcFace margin $m=0.1$, the weight for $L_{a}$ to $\beta=1$, and the weight for $L_r$ to $\lambda=10$.
The learnable scalar for the global loss $L_{g}$ is initialized to $\gamma=\sqrt{C_F}=45.25$.
See also the appendices for results obtained when training models on GLDv2 \cite{weyand2020google}.

\textbf{Evaluation datasets.} 
To evaluate our model, we use several datasets.
First, Oxford \cite{Philbin07} and Paris \cite{Philbin2008}, with revisited annotations \cite{radenovic2018revisiting}, referred to as $\mathcal{R}$Oxf and $\mathcal{R}$Par, respectively.
There are $4993$ ($6322$) database images in the $\mathcal{R}$Oxf ($\mathcal{R}$Par) dataset, and a different query set for each, both with $70$ images.
Performance is measured using mean average precision (mAP).
Large-scale results are further reported with the $\mathcal{R}$1M distractor set \cite{radenovic2018revisiting}, which contains $1$M images.
As in previous papers \cite{revaud2019learning,teichmann2019detect,ng2020solar}, parameters are tuned in $\mathcal{R}$Oxf/$\mathcal{R}$Par, then kept fixed for the large-scale experiments.
Second, we report large-scale instance-level retrieval and recognition results on the Google Landmarks dataset v2 (GLDv2) \cite{weyand2020google}, using the latest ground-truth version ($2.1$).
GLDv2-retrieval has $1129$ queries ($379$ validation and $750$ testing) and $762$k database images; performance is measured using mAP@$100$.
GLDv2-recognition has $118$k test ($41$k validation and $77$k testing) and $4$M training images from $203$k landmarks; the training images are only used to retrieve images and their scores/labels are used to form the class prediction; performance is measured using $\mu$AP@$1$.
We perform minimal parameter tuning based on the validation split, and report results on the testing split.

\textbf{Feature extraction and matching.}
We follow the convention from previous work \cite{radenovic2018fine,gordo2017end,noh2017large} and use an image pyramid at inference time to produce multi-scale representations.
For global features, we use $3$ scales, $\{\frac{1}{\sqrt{2}}, 1, \sqrt{2}\}$; $L_2$ normalization is applied for each scale independently, then the three global features are average-pooled, followed by another $L_2$ normalization step.
For local features, we experiment with the same $3$ scales, but also with the more expensive setting from \cite{noh2017large} using $7$ image scales in total, with range from $0.25$ to $2.0$ (this latter setting is used unless otherwise noted).
Local features are selected based on their attention scores $\mathcal{A}$; a maximum of $1$k local features are allowed, with a minimum attention score $\tau$, where we set $\tau$ to the median attention score in the last iteration of training, unless otherwise noted.
For local feature matching, we use RANSAC \cite{Fischler1981} with an affine model.
When re-ranking global feature retrieval results with local feature-based matching, the top $100$ ranked images from the first stage are considered.
For retrieval datasets, the final ranking is based on the number of inliers, then breaking ties using the global feature distance.
For the recognition dataset, we follow a similar protocol as \cite{weyand2020google} to produce class predictions by aggregating scores of top retrieved images, where the scores of top images are given by $\frac{min(i,70)}{70} + \alpha c$ (here, $i$ is the number of inliers, $c$ the global descriptor cosine similarity and $\alpha=0.25$).
Our focus is on improving image features for retrieval/recognition, so we do not consider techniques that post-process results such as query expansion \cite{chum2007total,radenovic2018fine} or diffusion/graph traversal \cite{iscen2017efficient,chang2019explore}.
These are expensive due to requiring additional passes over the database, but if desired could be integrated to our system and produce stronger performance.

\subsection{Results}

First, we present ablation experiments, to compare features produced by our joint model against their counterparts which are separately trained, and also to discuss the effect of controlling the gradient propagation.
For a fair comparison, our jointly trained features are evaluated against equivalent separately-trained models, with the same hyperparameters as much as possible.
Then, we compare our models against state-of-the-art techniques.
See also the appendices for more details, visualizations and discussions.


\textbf{Local features.}
As an ablation, we evaluate our local features by matching image pairs.
We select $200$k pairs, each composed of a test and a train image from GLD, where in $1$k pairs both images depict the same landmark, and in $199$k pairs the two images depict different landmarks.
We compute average precision (AP) after ranking the pairs based on the number of inliers.
All variants for this experiment use $\tau$ equals to the $75^{th}$ percentile attention score in the last iteration of training.
Results are presented in \tabref{tab:dr_ablation}.

\input{sections/tab_dr.tex}
\input{sections/tab_ablation.tex}

First, we train solely the attention and dimensionality reduction modules, for $500$k iterations, all methods initialized with the same weights from a separately-trained global feature model.
These results are marked as not being jointly trained.
It can be seen that our AE outperforms PCA and a simpler method using only a single fully-connected (FC) layer.
Performance improves for the AE as $\lambda$ increases from $0$ to $10$, decreasing with $20$.
Then, we jointly train the unified model; in this case, the variant that does not stop gradients to the backbone suffers a large drop in performance, while the variant that stops gradients obtains similar results as in the separately-trained case.

The poor performance of the naive jointly trained model is due to the degradation of the hierarchical feature representation.
This can be assessed by observing the evolution of activation sparsity in $\mathcal{S}$ (\textit{conv4}) and $\mathcal{D}$ (\textit{conv5}), as shown in \figref{fig:sparsity}.
Generally, layers representing more abstract and high-level semantic properties (usually deeper layers) have high levels of sparsity, while shallower layers representing low-level and more localizable patterns are dense.
As a reference, the ImageNet pre-trained model presents on average $45\%$ and $82\%$ sparsity for these two feature maps, respectively, when run over GLD images.
For the naive joint training case, the activations of both layers quickly become much sparser, reaching $80\%$ and $97\%$ at the end of training; in comparison, our proposed training scheme preserves similar sparsity as the ImageNet model: $45\%$ and $88\%$.
This suggests that the \textit{conv4} features in the naive case degrade for the purposes of local feature matching; controlling the gradient effectively resolves this issue.

\textbf{Global features.}
\tabref{tab:global_ablation} compares global feature training methods.
The first three rows present global features trained with different loss and pooling techniques.
We experiment with standard Softmax Cross-Entropy and ArcFace \cite{deng2019arcface} losses; for pooling, we consider standard average pooling (equivalent to SPoC \cite{babenko2015iccv}) and GeM \cite{radenovic2018fine}.
ArcFace brings an improvement of up to $14$\%, and GeM of up to $9.5$\%.
GeM pooling and ArcFace loss are adopted in our final model.
Naively training a joint model, without controlling gradients, underperforms when compared to the baseline separately-trained global feature, with mAP decrease of up to $5.7\%$.
Once gradient stopping is employed, the performance can be recovered to be on par with the separately-trained version (a little better on $\mathcal{R}$Oxf, a little worse on $\mathcal{R}$Par).
This is expected, since the global feature in this case is optimized by itself, without influence from the local feature head.

\textbf{Comparison to retrieval state-of-the-art.}
\tabref{tab:sota} compares our model against the retrieval state-of-the-art.
Three settings are presented: (A) local feature aggregation and re-ranking (previous work); (B) global feature similarity search; (C) global feature search followed by re-ranking with local feature matching and spatial verification (SP).

\input{sections/tab_sota.tex}

In setting (B), the DELG global feature variants strongly outperform previous work for all cases (most noticeably in the large-scale setting): $7.1\%$ absolute improvement in \roxf{}+1M-Hard and $7.5\%$ in \rpar{}+1M-Hard.
Note that we obtain strong improvements even when using the ResNet-50 backbone, while the previous state-of-the-art used ResNet-101/152, which are much more complex ($2$X/$3$X the number of floating point operations, respectively).
To ensure a fair comparison, we present results from \cite{radenovic2018fine,revaud2019learning} which specifically use the same training set as ours, marked as ``(GLD)'' -- the results are obtained from the authors' official codebases.
In particular, note that ``R152-GeM (GLD) \cite{radenovic2018fine}'' uses not only the same training set, but also the same exact scales in the image pyramid; even if our method is much cheaper, it consistently outperforms others.

For setup (C), we use both global and local features.
For large-scale databases, it may be impractical to store all raw local features in memory; to alleviate such requirement, we also present a variant, DELG$^\star$, where we store local features in binarized format, by simply applying an elementwise function: $b(x)=+1 \, \, \textrm{if}  \, \, x > 0, -1 \, \, \textrm{otherwise}$.

Local feature re-ranking boosts performance substantially for DELG, compared to only searching with global features, especially in large-scale cases: gains of up to $9\%$ (in \roxf{}+1M-Hard).
We also present results where local feature extraction is performed with $3$ scales only, the same ones used for global features.
The large-scale results are similar, providing a boost of up to $7.7\%$.
Results for DELG$^\star$ also provide large improvements, but with performance that is slightly lower than the corresponding unbinarized versions.
Our retrieval results also outperform DSM \cite{simeoni2019local} significantly, by more than $10\%$ in several cases.
Different from our proposed technique, the gain from spatial verification reported in their work is small, of at most $1.5\%$ absolute.
DELG also outperforms local feature aggregation results from setup (A) in $7$ out of $8$ cases, establishing a new state-of-the-art across the board.


%

\input{sections/tab_gld_and_reranking.tex}

\input{sections/tab_cost.tex}

\textbf{GLDv2 evaluation.}
\tabref{tab:gldv2_results} compares DELG against previous GLDv2 results, where for a fair comparison we report methods trained on GLD.
DELG achieves top performance in both retrieval and recognition tasks, with local feature re-ranking providing a significant boost in both cases -- especially on the recognition task ($29.2\%$ absolute improvement).
Note that recent work has reported even higher performance on the retrieval task, by learning on GLDv2's training set and using query expansion techniques \cite{yokoo2020two} / ensembling \cite{weyand2020google}.
On the other hand, DELG's performance on the recognition task is so far the best reported single-model result, outperforming many ensemble-based methods (by itself, it would have been ranked top-3 in the 2019 challenge) \cite{weyand2020google}.

\textbf{Re-ranking experiment.}
\tabref{tab:reranking} further compares local features for re-ranking purposes.
R50-DELG is compared against SIFT \cite{Lowe2004}, SOSNet \cite{tian2019sosnet} (HPatches model, DoG keypoints) and D2-Net \cite{Dusmanu2019CVPR} (trained, multiscale).
All methods are given the same retrieval short list of $100$ images for re-ranking (based on R50-DELG-global retrieval); for a fair comparison, all methods use $1$k features and $1$k RANSAC iterations.
We tuned matching parameters separately for each method: whether to use ratio test or distance threshold for selecting correspondences (and their associated thresholds); RANSAC residual threshold; minimum number of inliers (below which we declare no match).
SIFT and SOSNet provide little improvement over the global feature, due to suboptimal feature detection based on our observation (\ie, any blob-like feature is detected, which may not correspond to landmarks). D2-Net improves over the global feature, benefiting from a better feature detector. DELG outperforms other methods by a large margin. 

\textbf{Latency and memory.}
\tabref{tab:cost} reports feature extraction latency and index memory footprint for state-of-the-art methods, corresponding to the three settings from \tabref{tab:sota}; as a reference, we also present numbers for DELF (which uses an R50 backbone).
Joint extraction with DELG allows for substantial speed-up, compared to running two separate local and global models: when using $3$ local feature scales, separately running R50-GeM and DELF would lead to $198$ms, while the unified model runs with latency of $118$ms ($40\%$ faster).
For the R50 case with $7$ local scales, the unified model is $30\%$ faster.
The binarization technique adds negligible overhead, having roughly the same latency.

Storing unquantized DELG local features requires excessive index memory requirements; using binarization, this can be reduced significantly: R101-DELG$^\star$ requires $23$GB.
This is lower than the memory footprint of DELF-R-ASMK$^\star$.
Note also that feature extraction for DELG$^\star$ is much faster than for DELF-R-ASMK$^\star$, by more than $5\times$.
R50-DELG with $3$ scales is also faster than using a heavier global feature (R101-GeM \cite{radenovic2018fine}), besides being more accurate.
As a matter of fact, several of the recently-proposed global features \cite{radenovic2018fine,gordo2017end} use image pyramids with $3$ scales; our results indicate that their performance can be improved substantially by adding a local feature head, with small increase in extraction latency, and without degrading the global feature.

\textbf{Comparison against concurrent work.}
Our global feature outperforms the SOLAR \cite{ng2020solar} global feature method for all datasets and evaluation protocols, with gains of up to $3.3\%$.
Compared to Tolias \etal{} \cite{tolias2020learning}, we obtain similar results on \roxf{}/\rpar{}; for the large-scale setting, their best average mAP across the two datasets is $50.1\%$, while R101-DELG obtains $50.4\%$ and R101-DELG$^\star$ obtains $50.0\%$ (in general, our method is a little better on \rpar{} and theirs on \roxf{}).
On GLDv2-recognition, Tolias \etal{} \cite{tolias2020learning} achieve $36.5\%$ $\mu$AP, while R101-DELG obtains $61.2\%$.

\begin{figure*}[t]
\centering
\begin{subfigure}{0.5\textwidth}
  \centering
 \includegraphics[width=0.94\linewidth]{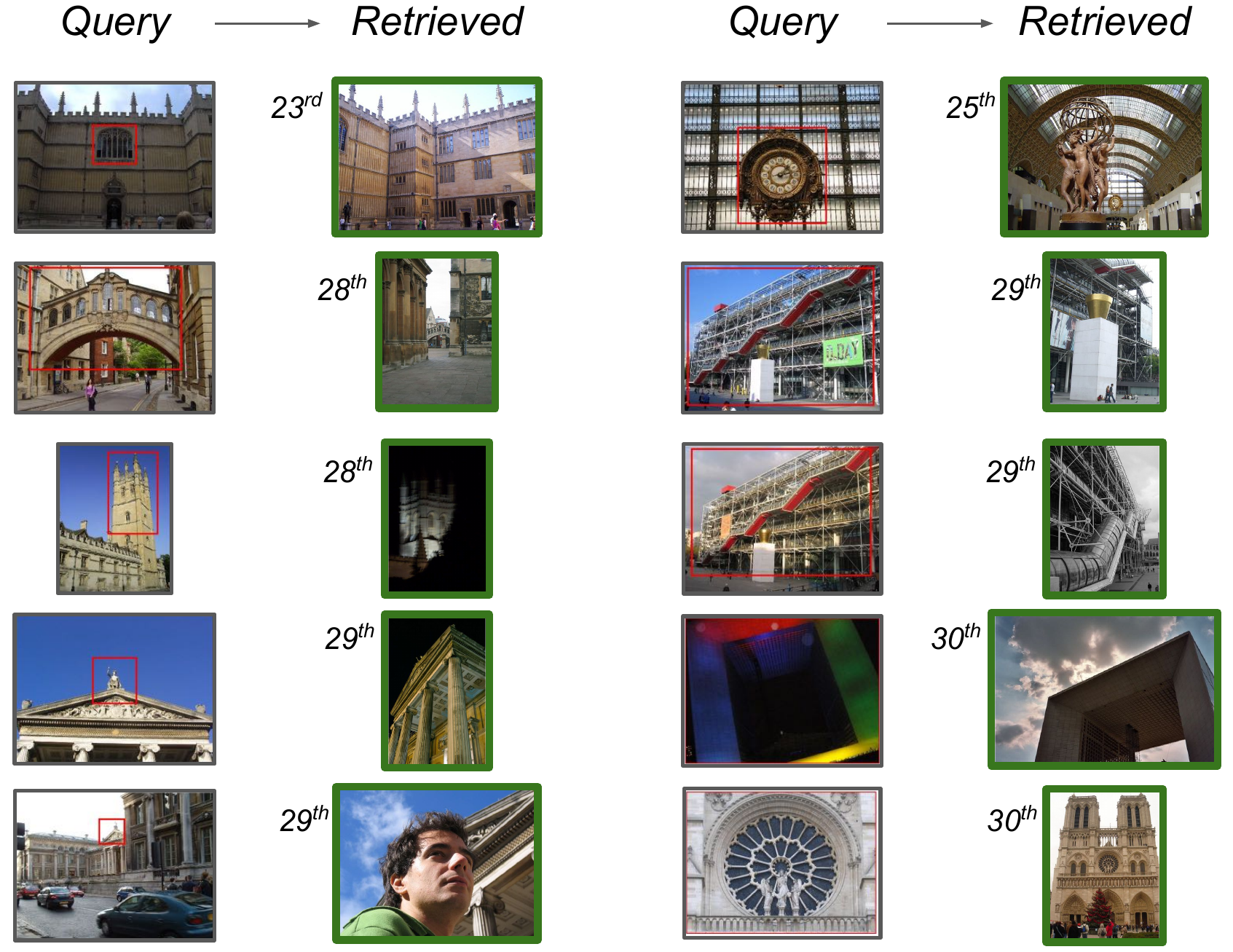}
  \caption{\textbf{Global feature retrieval: high recall.}}
  \label{fig:global_better}
\end{subfigure}%
\begin{subfigure}{0.5\textwidth}
  \centering
   \includegraphics[width=0.96\linewidth]{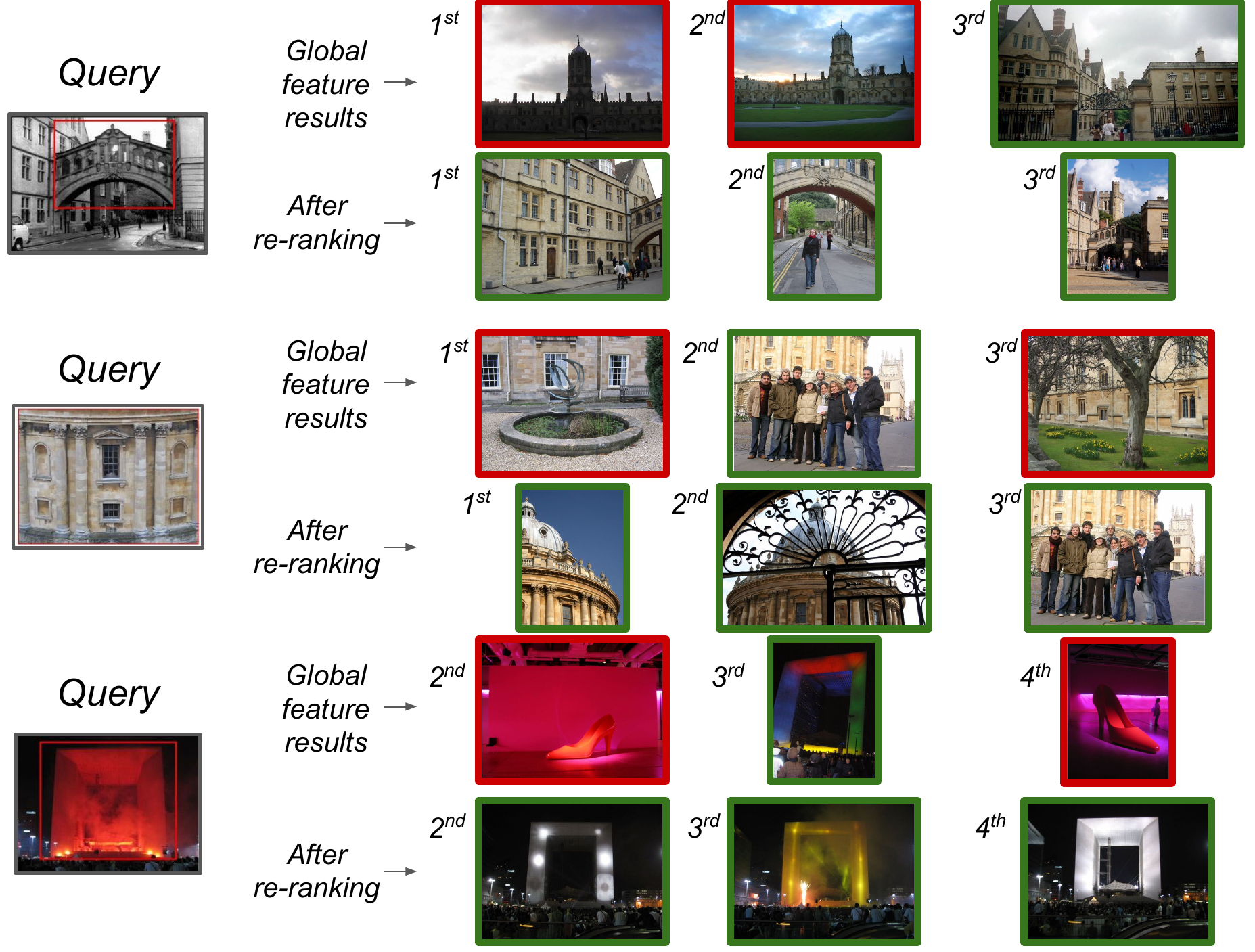}
  \caption{\textbf{Local feature re-ranking: high precision.}}
  \label{fig:local_better}
\end{subfigure}
\caption{Sample DELG results on \roxf{}-Hard and \rpar{}-Hard. \textbf{(a)} Examples of difficult, high-ranked relevant retrieved images for $10$ different queries.
These retrieved database images have a low number of inliers after geometric verification (if any), which means that their similarity is mainly captured by the global feature.
\textbf{(b)} Examples illustrating performance improvements from the re-ranking stage using local features.
For each query (left), two rows are presented on the right, the top one showing results based on global feature similarity and the bottom one showing results after re-ranking the top $100$ images with local features.
Correct results are marked with green borders, and incorrect ones in red.
While top retrieved global feature results are often ranked incorrectly, local feature matching can effectively re-rank them to improve precision.}
\label{fig:qualitative_global_and_local}
\end{figure*}

\textbf{Qualitative results.}
We give examples of retrieval results, to showcase the DELG model.
\figref{fig:global_better} illustrates difficult cases, where the database image shows a very different viewpoint, or significant lighting changes; these images can still achieve relatively high ranks due to effective global features, which capture well the similarity even in such challenging scenarios.
In these cases, local features do not produce sufficient matches.

\figref{fig:local_better} shows the effect of local feature re-ranking for selected queries, for which substantial gains are obtained.
Global features tend to retrieve images that have generally similar appearance, but which sometimes do not depict the same object of interest; this can be improved substantially with local feature re-ranking, which enables stricter matching selectivity.
As can be observed in these two figures, global features are crucial for high recall, while local features are key to high precision.

%% file: sections/tab_dr.tex

\newfloatcommand{capbtabbox}{table}[][\FBwidth]

\begin{figure}[t]
\begin{floatrow}
\capbtabbox{%
\scriptsize
\addtolength{\tabcolsep}{0.3em}
\centering
{
\begin{tabular}{c c c c c}
  \multirow{2}{*}{DR Method} & \multirow{2}{*}{$\lambda$}&Jointly& Stop & GLD-pairs \\
 &&trained&gradients& AP (\%) \\
\toprule
PCA \cite{noh2017large} &-&\multirow{2}{*}{\xmark}&\multirow{2}{*}{-}& $51.48$ \\
FC &-&&& $52.67$\\
\midrule
 \multirow{7}{*}{AE [ours]} &  $0$ &\multirow{5}{*}{\xmark} &\multirow{5}{*}{-}& $49.95$ \\ 
 & \multirow{1}{*}{$1$}&&  & $51.28$ \\
 & \multirow{1}{*}{$5$}&&  & $52.26$ \\
 & \multirow{1}{*}{$10$}&& & $54.21$ \\
 & \multirow{1}{*}{$20$}&& & $53.51$ \\ \cmidrule[0.5pt]{2-5}
  & 10&\multirow{2}{*}{\cmark}&\xmark & $37.05$ \\
  & 10&&\cmark & $53.73$ \\
  \\
  \\
  \\
\end{tabular}
}
}
{%
  \caption{\textbf{Local feature ablation.} Comparison of local features, trained separately or jointly, with different methods for dimensionality reduction (DR). We report average precision (AP) results of matching image pairs from the Google Landmarks dataset (GLD).}%
  \label{tab:dr_ablation}
}
\ffigbox{%
\begin{center}
   \includegraphics[width=0.95\linewidth]{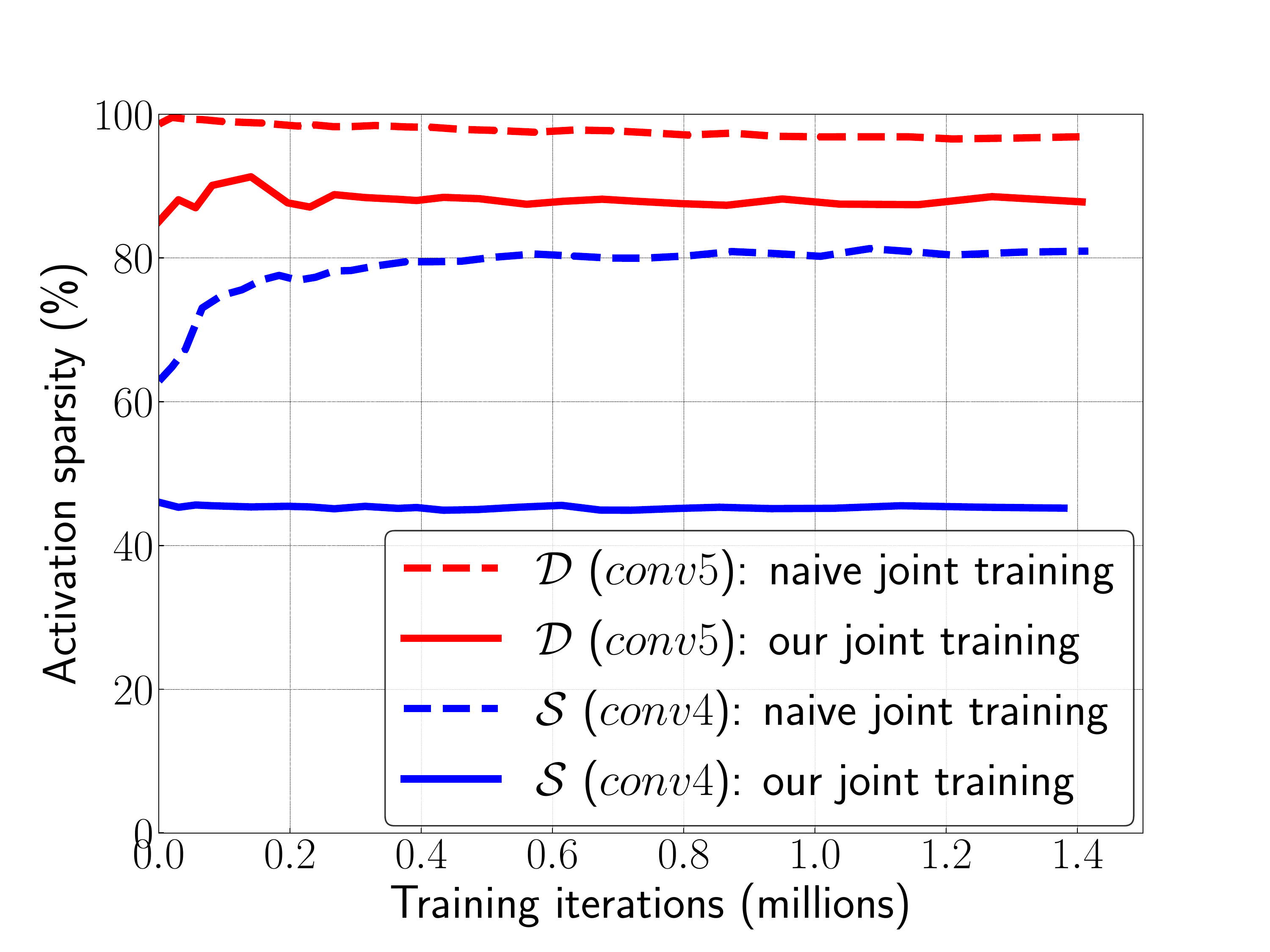}
\end{center}
}{%
   \caption{Evolution of \textbf{activation sparsity} over training iterations for $\mathcal{D}$ (\textit{conv5}) and $\mathcal{S}$ (\textit{conv4}), comparing the naive joint training method and our improved version that controls gradient propagation. The naive method leads to much sparser feature maps.}
   \label{fig:sparsity}
}
\end{floatrow}
\end{figure}

%% file: sections/tab_ablation.tex

\begin{table*}[t]
\scriptsize
\addtolength{\tabcolsep}{0.6em}
\centering
{
\begin{tabular}{c c c c c c c c c c}
&&Jointly & Stop && \multicolumn{2}{c}{ \scriptsize Medium} && \multicolumn{2}{c}{ \scriptsize Hard} \\
Pooling & Loss & trained  & gradients && \multicolumn{1}{c}{\scriptsize $\mathcal{R}$Oxf} & \multicolumn{1}{c}{\scriptsize $\mathcal{R}$Par} && \multicolumn{1}{c}{\scriptsize $\mathcal{R}$Oxf} & \multicolumn{1}{c}{\scriptsize $\mathcal{R}$Par} \\
\midrule
SPoC & Softmax & \xmark & --  && $51.2$ & $72.0$ && $26.3$ & $47.7$ \\
SPoC & ArcFace & \xmark & -- && $59.8$ & $80.8$ && $35.6$ & $61.7$ \\
GeM & ArcFace & \xmark & -- && $69.3$ & \num[math-rm=\mathbf]{82.2} && $44.4$ & \num[math-rm=\mathbf]{64.0}\\
GeM & ArcFace & \cmark & \xmark && $68.8$ & $78.9$ && $42.4$ & $58.3$ \\
GeM & ArcFace & \cmark & \cmark && \num[math-rm=\mathbf]{69.7} & $81.6$ && \num[math-rm=\mathbf]{45.1} & $63.4$ \\
\end{tabular}
}
\caption{\textbf{Global feature ablation.}
Comparison of global features, trained separately or jointly, with different pooling methods (SPoC, GeM) and loss functions (Softmax, ArcFace). We report mean average precision (mAP \%) on the \roxf{} and \rpar{} datasets.}
\label{tab:global_ablation}
\end{table*}

%% file: sections/tab_sota.tex

\begin{table*}[t]
\scriptsize
\addtolength{\tabcolsep}{0.32em}
\centering
{
\begin{tabular}{l c c c c c c c c c }
 \small \multirow{3}{*}{Method}  & \multicolumn{4}{c}{ \small Medium} && \multicolumn{4}{c}{ \small Hard} \\[+0.3em] \cmidrule[0.5pt]{2-5} \cmidrule[0.5pt]{7-10}
 & \multicolumn{1}{c}{\large \vphantom{M} \scriptsize $\mathcal{R}$Oxf } & \multicolumn{1}{c}{\scriptsize +1M} & \multicolumn{1}{c}{\scriptsize $\mathcal{R}$Par} & \multicolumn{1}{c}{\scriptsize +1M} && \multicolumn{1}{c}{\scriptsize $\mathcal{R}$Oxf} & \multicolumn{1}{c}{\scriptsize +1M} & \multicolumn{1}{c}{\scriptsize $\mathcal{R}$Par} & \multicolumn{1}{c}{\scriptsize +1M} \\
\midrule
{ \textit{(A) Local feature aggregation + re-ranking}} \\ 
HesAff-rSIFT-ASMK$^\star$+SP \cite{tolias2015image} &  \num{60.6} & \num{46.8} & \num{61.4} & \num{42.3} &&  \num{36.7} & \num{26.9} & \num{35.0} & \num{16.8} \\ 
HesAff-HardNet-ASMK$^\star$+SP \cite{mishkin2018repeatability} &  \num{65.6}  & -- & \num{65.2} & -- &&  \num{41.1} & -- & \num{38.5} & -- \\ 
DELF-ASMK$^\star$+SP \cite{noh2017large,radenovic2018revisiting} & \num{67.8}  & \num{53.8} & \num{76.9} & \num{57.3} && \num{43.1} & \num{31.2} & \num{55.4} & \num{26.4} \\ 
DELF-R-ASMK$^\star$+SP (GLD) \cite{teichmann2019detect}  & \num[math-rm=\mathbf]{76.0}  & \num[math-rm=\mathbf]{64.0} & \num[math-rm=\mathbf]{80.2} & \num[math-rm=\mathbf]{59.7} && \num[math-rm=\mathbf]{52.4} & \num[math-rm=\mathbf]{38.1} & \num[math-rm=\mathbf]{58.6} & \num[math-rm=\mathbf]{29.4} \\
\midrule
{ \textit{(B) Global features}} \\
AlexNet-GeM \cite{radenovic2018fine} &   \num{43.3} & \num{24.2}  & \num{58.0}  & \num{29.9} &&  \num{17.1} & \num{9.4} & \num{29.7} & \num{8.4} \\
VGG16-GeM \cite{radenovic2018fine} &     \num{61.9} &  \num{42.6} & \num{69.3}  & \num{45.4} && \num{33.7} & \num{19.0} & \num{44.3} & \num{19.1} \\ 
R101-R-MAC \cite{gordo2017end} &   \num{60.9}  & \num{39.3} & \num{78.9} & \num{54.8} && \num{32.4} & \num{12.5} & \num{59.4} & \num{28.0} \\ 
R101-GeM \cite{radenovic2018fine} & \num{64.7}  & \num{45.2} & \num{77.2}  & \num{52.3} && \num{38.5} & \num{19.9} & \num{56.3} & \num{24.7} \\
R101-GeM$\uparrow$ \cite{simeoni2019local} & \num{65.3} & \num{46.1} & \num{77.3} & \num{52.6} &&  \num{39.6} & \num{22.2} & \num{56.6} & \num{24.8} \\
R101-GeM-AP \cite{revaud2019learning} & \num{67.5} & \num{47.5} & \num{80.1} & \num{52.5} && \num{42.8} & \num{23.2} & \num{60.5} & \num{25.1} \\
R101-GeM-AP (GLD) \cite{revaud2019learning} & \num{66.3} & -- & \num{80.2} & -- && \num{42.5} & -- & \num{60.8} & -- \\
R152-GeM (GLD) \cite{radenovic2018fine} & \num{68.7}  & -- & \num{79.7}  & -- && \num{44.2} & -- & \num{60.3} & -- \\
R50-DELG \textbf{[ours]} & $69.7$ & \num[math-rm=\mathbf]{55.0} & $81.6$ & $59.7$ && $45.1$ &$27.8$ & $63.4$ & $34.1$ \\ 
R101-DELG \textbf{[ours]} & \num[math-rm=\mathbf]{73.2} & $54.8$ & \num[math-rm=\mathbf]{82.4} & \num[math-rm=\mathbf]{61.8} && \num[math-rm=\mathbf]{51.2} & \num[math-rm=\mathbf]{30.3} & \num[math-rm=\mathbf]{64.7} & \num[math-rm=\mathbf]{35.5} \\ 
\midrule
{ \textit{(C) Global features + Local feature re-ranking} }\\
R101-GeM$\uparrow$+DSM \cite{simeoni2019local} & \num{65.3} & \num{47.6} & \num{77.4} & \num{52.8} &&  \num{39.2} & \num{23.2} & \num{56.2} & \num{25.0} \\
R50-DELG \hspace{2.5pt} \textbf{[ours]} & $75.1$ &$61.1$ & $82.3$ & $60.5$ && $54.2$ & $36.8$ & $64.9$ & $34.8$ \\ 
R50-DELG$^\star$ \textbf{[ours]} & -- & $60.4$ & -- & $60.3$ && -- & $35.3$ & -- & $34.1$ \\ 
R101-DELG \hspace{2.5pt} (3 scales global \& local) \textbf{[ours]} & $77.5$ & $61.7$ & \num[math-rm=\mathbf]{83.0} & $62.6$  && $56.7$ & $38.0$ & $65.2$ & $36.2$ \\ 
R101-DELG$^\star$ (3 scales global \& local) \textbf{[ours]} & -- & $61.6$ & -- & $62.3$  && -- & $36.9$ & -- & $35.5$ \\ 
R101-DELG \hspace{2.5pt} \textbf{[ours]} & \num[math-rm=\mathbf]{78.5} & \num[math-rm=\mathbf]{62.7} & $82.9$ & \num[math-rm=\mathbf]{62.6} && \num[math-rm=\mathbf]{59.3} & \num[math-rm=\mathbf]{39.3} & \num[math-rm=\mathbf]{65.5} & \num[math-rm=\mathbf]{37.0} \\ 
R101-DELG$^\star$ \textbf{[ours]} & -- & $62.5$ & -- & $62.5$ && -- & $38.5$ & -- & $36.3$ \\ 
\end{tabular}
}
\caption{\textbf{Comparison to retrieval state-of-the-art.} Results (\% mAP) on the $\mathcal{R}$Oxf/$\mathcal{R}$Par datasets (and their large-scale versions $\mathcal{R}$Oxf+1M/$\mathcal{R}$Par+1M), with both Medium and Hard evaluation protocols.
The top set of rows (A) presents previous work's results using local feature aggregation and re-ranking.
Other sets of rows present results using (B) global features only, or (C) global features for initial search then re-ranking using local features.
DELG$^\star$ refers to a version of DELG where the local features are binarized.
DELG and DELG$^\star$ outperform previous work in setups (B) and (C) substantially.
DELG also outperforms methods from setting (A) in $7$ out of $8$ cases.}
\label{tab:sota}
\end{table*}

%% file: sections/tab_gld_and_reranking.tex

\begin{figure}[t]
\begin{floatrow}
\capbtabbox{%
\scriptsize
\addtolength{\tabcolsep}{0.3em}
\centering
{
\begin{tabular}{l c c}
  \multirow{2}{*}{Method} & Retrieval & Recognition \\
&mAP (\%)& $\mu$AP (\%) \\
\toprule
DELF-R-ASMK*+SP \cite{teichmann2019detect} & $18.8$ & -- \\
R101-GeM+ArcFace \cite{weyand2020google} & $20.7$ & $33.3$ \\ 
R101-GeM+CosFace \cite{yokoo2020two} & $21.4$ & -- \\ 
DELF-KD-tree \cite{noh2017large} &-- & $44.8$ \\
\midrule
R50-DELG (global-only) [ours]  & $20.4$ & $32.4$ \\
R101-DELG (global-only) [ours]& $21.7$ & $32.0$ \\
R50-DELG [ours]  & $22.3$ & $59.2$ \\
R101-DELG [ours] & \num[math-rm=\mathbf]{24.5} & \num[math-rm=\mathbf]{61.2} \\
  \\
\end{tabular}
}
}
{%
  \caption{\textbf{GLDv2 evaluation.} Results on the GLDv2 dataset, for the retrieval and recognition tasks, on the ``testing" split of the query set. For a fair comparison, all methods are trained on GLD.}%
  \label{tab:gldv2_results}
}
\capbtabbox{%
\scriptsize
\addtolength{\tabcolsep}{0.3em}
\centering
\begin{tabular}{l c c}
  Method & Hard & Medium \\
\toprule
R50-DELG (global-only)  & $45.1$ & $69.7$ \\
\midrule
{ \textit{Local feature re-ranking} }\\
SIFT \cite{Lowe2004} & $44.4$ & $69.8$ \\
SOSNet \cite{tian2019sosnet} & $45.5$ & $69.9$ \\
D2-Net \cite{Dusmanu2019CVPR} & $47.2$ & $70.4$ \\
R50-DELG [ours] & \num[math-rm=\mathbf]{54.2} & \num[math-rm=\mathbf]{75.1} \\
  \\
  \\
  \\
\end{tabular}
}
{%
  \caption{\textbf{Re-ranking experiment.} Comparison of DELG against other recent local features; results (\% mAP) on the \roxf{} dataset.}%
  \label{tab:reranking}
}
\end{floatrow}
\end{figure}

%% file: sections/tab_cost.tex

\begin{table*}[t]
\scriptsize
\addtolength{\tabcolsep}{0.4em}
\centering
{
\begin{tabular}{l c c c}
&Extraction& \multicolumn{2}{c}{ \scriptsize Memory (GB)} \\
Method & latency (ms) & $\mathcal{R}$Oxf+1M & $\mathcal{R}$Par+1M \\
\toprule
\textit{(A) Local feature aggregation} \\
DELF-R-ASMK$^\star$ \cite{teichmann2019detect} & $2260$ & $27.6$ & -- \\
\midrule
\textit{(B) Global features} \\
R50-GeM \cite{radenovic2018fine} & $100$ & $7.7$ & $7.7$ \\
R101-GeM \cite{radenovic2018fine} & $175$ & $7.7$ & 7.7 \\
\midrule
\textit{(C) Unified global + local features} \\
R50-DELG \hspace{2.5pt} (3 scales global \& local) [ours] & $118$ & $439.4$ & $440.0$ \\
R50-DELG \hspace{2.5pt} [ours] & $211$ & $485.5$ & $486.2$ \\
R101-DELG \hspace{2.5pt} (3 scales global \& local) [ours] & $193$ & $437.1$ &  $437.8$ \\
R101-DELG$^\star$ (3 scales global \& local) [ours] & $193$ & $21.1$ & $21.1$ \\
R101-DELG \hspace{2.5pt} [ours] & $383$ & $485.9$ & $486.6$ \\
R101-DELG$^\star$ [ours] & $383$ & $22.6$ & $22.7$ \\
\midrule
\midrule
\textit{Local features} \\
DELF (3 scales) \cite{noh2017large} & $98$ & $434.2$ & $434.8$ \\
DELF (7 scales) \cite{noh2017large} & $201$ & $477.9$ & $478.5$ \\
\end{tabular}
}
\caption{Feature extraction \textbf{latency} and database \textbf{memory} requirements for different image retrieval models. Latency is measured on an NVIDIA Tesla P100 GPU, for square images of side $1024$. (A) DELF-R-ASMK$^\star$ measurements use the code and default configuration from \cite{teichmann2019detect}; (B) ResNet-GeM variants use $3$ image scales; (C) DELG and DELG$^\star$ are compared with different configurations. As a reference, we also provide numbers for DELF in the last rows.
}
\label{tab:cost}
\end{table*}

%% file: sections/conclusion.tex
\section{Conclusions}

Our main contribution is a unified model that enables joint extraction of local and global image features, referred to as DELG.
The model is based on a ResNet backbone, leveraging generalized mean pooling to produce global features and attention-based keypoint detection to produce local features.
We also introduce an effective dimensionality reduction technique that can be integrated into the same model, based on an autoencoder.
The entire network can be trained end-to-end using image-level labels and does not require any additional post-processing steps.
For best performance, we show that it is crucial to stop gradients from the attention and autoencoder branches into the network backbone, otherwise a suboptimal representation is obtained.
We demonstrate the effectiveness of our method with comprehensive experiments, achieving state-of-the-art performance on the Revisited Oxford, Revisited Paris and Google Landmarks v2 datasets.

%% file: sections/training_cost.tex

\section*{Appendix A. Training cost}

One of the advantages of DELG is that local and global features can be jointly trained in one shot, without the need of additional steps.
In practice, we have observed that the training time for DELG is roughly the same as for the associated global feature, both taking approximately $1.5$ day.

This is because the additional cost of learning the attention and autoencoder layers is small: there are only $5$ extra trainable layers ($2$ in the attention module, $2$ in the autoencoder, plus the attention loss classifier), and their gradients are not backpropagated to the network backbone.
Other factors play a much more significant role in the training speed, \eg{}: reading data from disk, transferring batch to GPU memory, applying image pre-processing operations such as resizing/cropping/augmentation, etc.

In short, training our local image features comes at a very small cost on top of global feature learning.
Let us clarify, though, that while the training cost is roughly the same between the global and joint models, the inference cost for the joint model has an overhead for local feature detection (\eg{}, selecting the top local features across different scales in the image pyramid).

We can also compare DELG's training cost to DELF's: DELF would require one additional run for attention learning ($6$ hours), followed by a PCA computation step ($3$ hours).
The advantage of DELG, compared to DELF, is that the attention and autoencoder layers are already adapting to the network backbone while it is training; for DELF, this process only happens after the backbone is fully trained.

%% file: sections/parameters.tex

\section*{Appendix B. Model selection and tuning}

We provide more details on how our models were selected/tuned, and specify chosen parameters which were not mentioned in the main text.
For more information, please refer to our released code/models.

\textbf{Model selection.}
Our models are trained for 1.5M steps, corresponding to approximately 25 epochs of an $80\%$ split of the GLD training set.
We attempt three different initial learning rates for each configuration, and select the best one based on the performance on \roxf{}/\rpar{} (it is a convention in recent work \cite{revaud2019learning,teichmann2019detect,ng2020solar} to perform ablations/tuning on these datasets).
In all cases, we pick the model at the end of training and do not hand-pick earlier checkpoints which may have higher performance.
These selected models are then used in all large-scale experiments, on \roxf{}+1M, \rpar{}+1M, GLDv2-retrieval and GLDv2-recognition.

\textbf{Tuning image matching.}
For local feature-based matching with DELG, we experimented with two methods for proposing putative correspondences (before feeding them to RANSAC): ratio test \cite{Lowe2004} and distance criteria.
RANSAC is used with $1$k iterations, and any returned match is used for re-ranking (minimum number of inliers is zero -- better results may be obtained by tuning the minimum number of inliers, which we did not do).
We tuned the local descriptor matching threshold (either ratio or distance threshold) and the RANSAC residual threshold; for the recognition task, we further tuned $\alpha$ (a weight used to combine the local and global scores -- see \secref{subsec:setup}).
Tuning is performed on \roxf{}/\rpar{}, and then the best configuration is fixed for experiments on \roxf{}+1M and \rpar{}+1M; similarly, we tune methods on the validation set of GLDv2-retrieval/GLDv2-recognition, then fix them for experiments on the testing set of GLDv2-retrieval/GLDv2-recognition.
For DELG on \roxf{}/\rpar{}, we use a ratio test with threshold $0.95$; for DELG$^\star$, we use distance-based matching with threshold $1.1$.
For DELG on GLDv2-retrieval/GLDv2-recognition, we also use distance-based matching, but in this case with a tighter threshold of $0.9$.
For all retrieval datasets, we set the RANSAC residual threshold to $20.0$; on the recognition dataset, it is set to $10.0$ (since the recognition task has a stronger focus on precision, we find that tighter matching parameters perform better).
For the recognition dataset, $\alpha$ was tuned to $0.25$.

%% file: sections/gldv2_trained_results.tex

\section*{Appendix C. Results with models trained on GLDv2}

\begin{table*}[t]
\scriptsize
\addtolength{\tabcolsep}{0.22em}
\centering
{
\begin{tabular}{l c c c c c c c c c c }
 \small \multirow{3}{*}{Method} & \multirow{3}{*}{\small GLD} & \multicolumn{4}{c}{ \small Medium} && \multicolumn{4}{c}{ \small Hard} \\[+0.3em] \cmidrule[0.5pt]{3-6} \cmidrule[0.5pt]{8-11}
& \small version & \multicolumn{1}{c}{\large \vphantom{M} \scriptsize $\mathcal{R}$Oxf } & \multicolumn{1}{c}{\scriptsize +1M} & \multicolumn{1}{c}{\scriptsize $\mathcal{R}$Par} & \multicolumn{1}{c}{\scriptsize +1M} && \multicolumn{1}{c}{\scriptsize $\mathcal{R}$Oxf} & \multicolumn{1}{c}{\scriptsize +1M} & \multicolumn{1}{c}{\scriptsize $\mathcal{R}$Par} & \multicolumn{1}{c}{\scriptsize +1M} \\
\midrule
{ \textit{Global features}} \\
R50-DELG & v1 & $69.7$ & $55.0$ & $81.6$ & $59.7$ && $45.1$ &$27.8$ & $63.4$ & $34.1$ \\ 
R50-DELG & v2-clean & $73.6$ & $60.6$ & $85.7$ & $68.6$ && $51.0$ & $32.7$ & $71.5$ & $44..4$ \\
R101-DELG &v1& $73.2$ & $54.8$ & $82.4$ & $61.8$ && $51.2$ & $30.3$ & $64.7$ & $35.5$ \\ 
R101-DELG & v2-clean & \num[math-rm=\mathbf]{76.3} & \num[math-rm=\mathbf]{63.7} & \num[math-rm=\mathbf]{86.6} & \num[math-rm=\mathbf]{70.6} && \num[math-rm=\mathbf]{55.6} & \num[math-rm=\mathbf]{37.5} & \num[math-rm=\mathbf]{72.4} & \num[math-rm=\mathbf]{46.9} \\
\midrule
{ \textit{Global features + Local feature re-ranking} }\\
R50-DELG & v1 & $75.4$ &$61.1$ & $82.3$ & $60.5$ && $54.2$ & $36.8$ & $64.9$ & $34.8$ \\ 
R50-DELG & v2-clean & $78.3$ & $67.2$ & $85.7$ & $69.6$ && $57.9$ & $43.6$ & $71.0$ & $45.7$ \\
R101-DELG \hspace{2.5pt} (3 scales global \& local) & v1 & $77.5$ & $61.7$ & $83.0$ & $62.6$  && $56.7$ & $38.0$ & $65.2$ & $36.2$ \\ 
R101-DELG \hspace{2.5pt} (3 scales global \& local) & v2-clean & $80.3$ & $68.6$ & $87.2$ & $71.4$ && $61.4$ & $45.3$ & $72.3$ & $47.7$ \\
R101-DELG$^\star$ (3 scales global \& local) & v1 & -- & $61.6$ & -- & $62.3$  && -- & $36.9$ & -- & $35.5$ \\ 
R101-DELG$^\star$ (3 scales global \& local) & v2-clean & -- & $67.4$ & -- & $71.0$ && -- & $42.9$ & -- & $46.5$ \\
R101-DELG & v1 & $78.5$ & $62.7$ & $82.9$ & $62.6$ && $59.3$ & $39.3$ & $65.5$ & $37.0$ \\ 
R101-DELG & v2-clean & \num[math-rm=\mathbf]{81.2} & \num[math-rm=\mathbf]{69.1} & \num[math-rm=\mathbf]{87.2} & \num[math-rm=\mathbf]{71.5} && \num[math-rm=\mathbf]{64.0} & \num[math-rm=\mathbf]{47.5} & \num[math-rm=\mathbf]{72.8} & \num[math-rm=\mathbf]{48.7} \\
R101-DELG$^\star$ & v1 & -- & $62.5$ & -- & $62.5$ && -- & $38.5$ & -- & $36.3$ \\ 
R101-DELG$^\star$ & v2-clean & -- & $68.6$ & -- & $71.1$ && -- & $45.1$ & -- & $47.3$ \\
\end{tabular}
}
\caption{\textbf{Training dataset comparison for Revisited Oxford/Paris evaluations.} Results (\% mAP) on the $\mathcal{R}$Oxf/$\mathcal{R}$Par datasets (and their large-scale versions $\mathcal{R}$Oxf+1M/$\mathcal{R}$Par+1M), with both Medium and Hard evaluation protocols, comparing models trained on GLD (v1) and GLDv2-clean. Training on GLDv2-clean provides a substantial performance boost in all cases.}
\label{tab:retrieval_gld_vs_gldv2}
\end{table*}

All results presented in the main part of the paper use models that were trained on the first version (v1) of the Google Landmarks Dataset, referred to as GLD \cite{noh2017large}.
In this appendix, we provide experimental results based on models that were trained on GLDv2 \cite{weyand2020google}.
Besides being a larger and more comprehensive dataset, GLDv2 has the advantage of stability -- all images have permissive licenses which allow indefinite retention; in contrast, GLD shrinks over time as images get deleted by users who uploaded them, due to copyright restrictions.
We believe that, moving forward, it may be more suitable to rely on stable datasets such as GLDv2 for proper experimental comparisons, since in this case one can guarantee that identical datasets are used for training and evaluation.

For our experiments, we used a subset of the entire training set, referred to as ``GLDv2-train-clean'' (GLDv2-clean for short), containing $1.6M$ images of $81k$ landmarks.
We divide it into two subsets `train'/`val' with $80\%$/$20\%$ split.
We train for $2$M steps (corresponding to approximately $25$ epochs of the `train' split).
All other training hyperparameters are kept the same as described in \secref{subsec:setup}.
When evaluating on retrieval/recognition datasets, we did not tune any evaluation-related parameters and simply used the configurations that worked best for the GLD-trained models.

Image retrieval results on the $\mathcal{R}$Oxf/$\mathcal{R}$Par datasets are presented on \tabref{tab:retrieval_gld_vs_gldv2}, showing consistent improvement in all cases.
First, consider results based solely on global features (top rows).
There are striking mAP improvements, especially for the large-scale evaluations: for example, $11.4\%$ absolute improvement on \rpar{}+1M-Hard and $8.9\%$ on \roxf{}+1M-Medium, both results with the R101 backbone.
Second, consider results based on global feature search followed by local feature re-ranking (bottom rows).
Again, we see substantial improvement: for example, R101-DELG absolute improvements of $11.7\%$ on \rpar{}+1M-Hard and $8.2\%$ on \roxf{}+1M-Hard.
In most large-scale cases, the improvement due to local feature re-ranking increases when using GLDv2-clean (with the exception of \roxf{}+1M-Medium), which may be due to better local features and/or a better short list of images selected for re-ranking.

Instance-level retrieval/recognition results on GLDv2 are presented on \tabref{tab:gldv2eval_gld_vs_gldv2}.
Global-only retrieval results are improved significantly, with $4.3\%$ absolute gain for R101-DELG; when re-ranking retrieval results with local feature-based geometric verification, the gains are smaller but still substantial, of $2.5\%$.
For the recognition task, we surprisingly observe a decrease in performance when training the model on GLDv2-clean: decrease of $3.1\%$ on the global-only setup, and $4.8\%$ when re-ranking with local features, both numbers again for the R101-based model.
Note that a similar observation was reported in \cite{weyand2020google}.

\begin{table*}[t]
\scriptsize
\addtolength{\tabcolsep}{0.3em}
\centering
\begin{tabular}{l c c c}
\multirow{2}{*}{Method} & GLD-train & Retrieval & Recognition \\
& version &mAP (\%)& $\mu$AP (\%) \\
\toprule
R50-DELG (global-only) & v1 & $20.4$ & $32.4$ \\
R50-DELG (global-only) & v2-clean & $24.1$ & $27.9$ \\
R101-DELG (global-only) & v1 & $21.7$ & $32.0$ \\
R101-DELG (global-only) & v2-clean & $26.0$ & $28.9$ \\
R50-DELG & v1 & $22.3$ & $59.2$ \\
R50-DELG & v2-clean & $24.3$ & $55.2$ \\
R101-DELG & v1 & $24.3$ & \num[math-rm=\mathbf]{61.2} \\
R101-DELG & v2-clean & \num[math-rm=\mathbf]{26.8} & $56.4$ \\
  \\
\end{tabular}
\caption{\textbf{Training dataset comparison for GLDv2 evaluations.} Results on the GLDv2 dataset, for the retrieval and recognition tasks, on the ``testing" split of the query set, comparing models trained on GLD (v1) and GLDv2-clean. Training on GLDv2-clean provides a performance boost for the retrieval task, but a small degradation for the recognition task.}
\label{tab:gldv2eval_gld_vs_gldv2}
\end{table*}

%% file: sections/memory.tex

\section*{Appendix D. Memory footprint}

For reporting the memory footprint of DELG/DELG$^\star$, we follow a similar convention to previous work \cite{radenovic2018revisiting,teichmann2019detect} and report total storage required for local and global descriptors.
Note that local feature geometry information is not counted in the reported numbers, and would add some overhead.

Our main focus in this paper is to propose a new model for unified local and global feature extraction, so we did not thoroughly explore techniques for efficient quantization.
As the DELG$^\star$ results show, there is great promise in aggressively quantizing local descriptors, leading to reasonable storage requirements -- which can likely be improved with more effective quantizers.
Similarly, global descriptors and local feature geometry could be quantized to improve the total memory cost substantially.

%% file: sections/viz_results.tex

\section*{Appendix E. Local feature matching visualizations}

\begin{figure*}
\centering
\includegraphics[width=\textwidth]{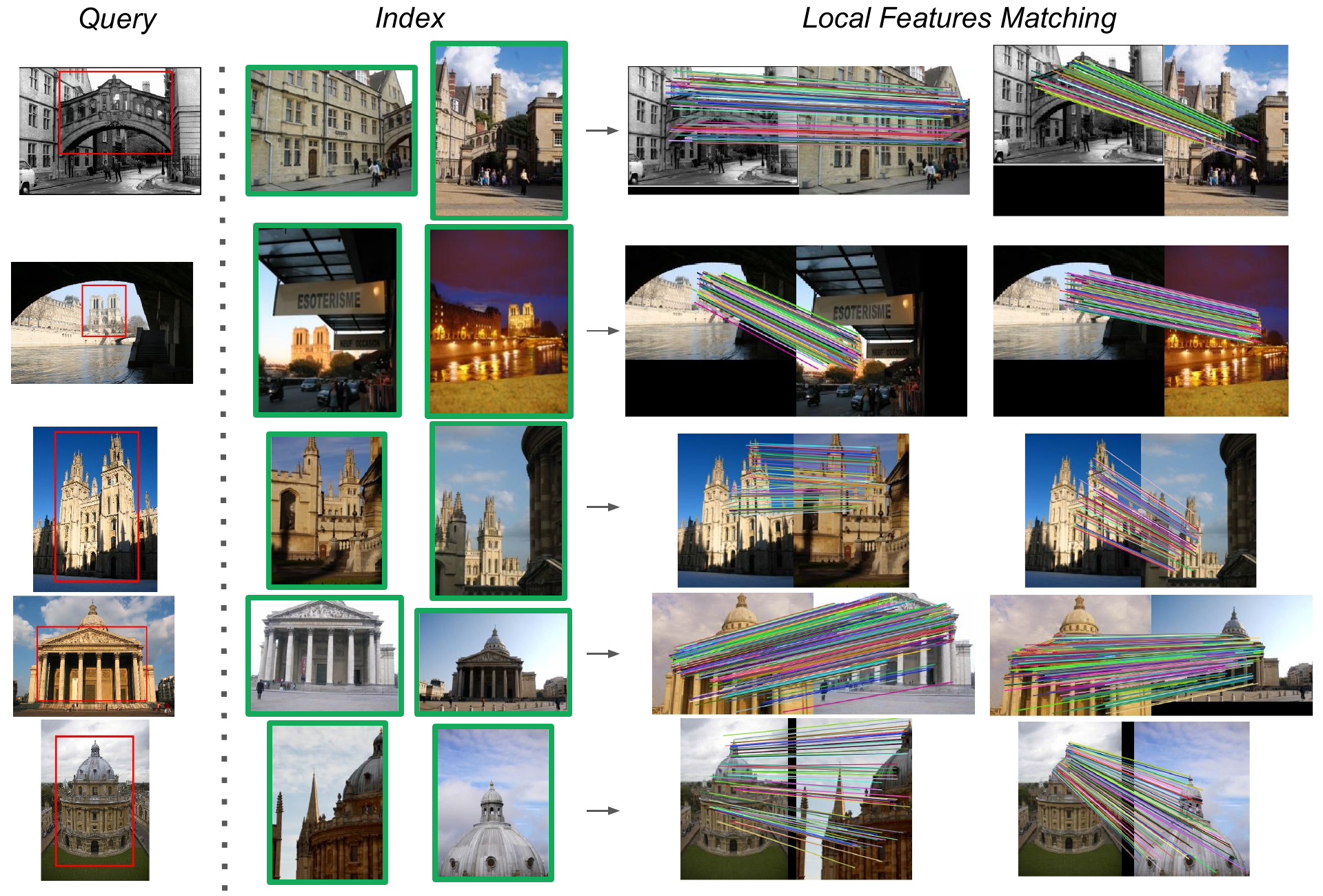}
\caption{Examples of correct local feature matches, for image pairs depicting the same object/scene.}
\label{fig:positive_matching}
\end{figure*}

\begin{figure*}
\centering
\includegraphics[width=\textwidth]{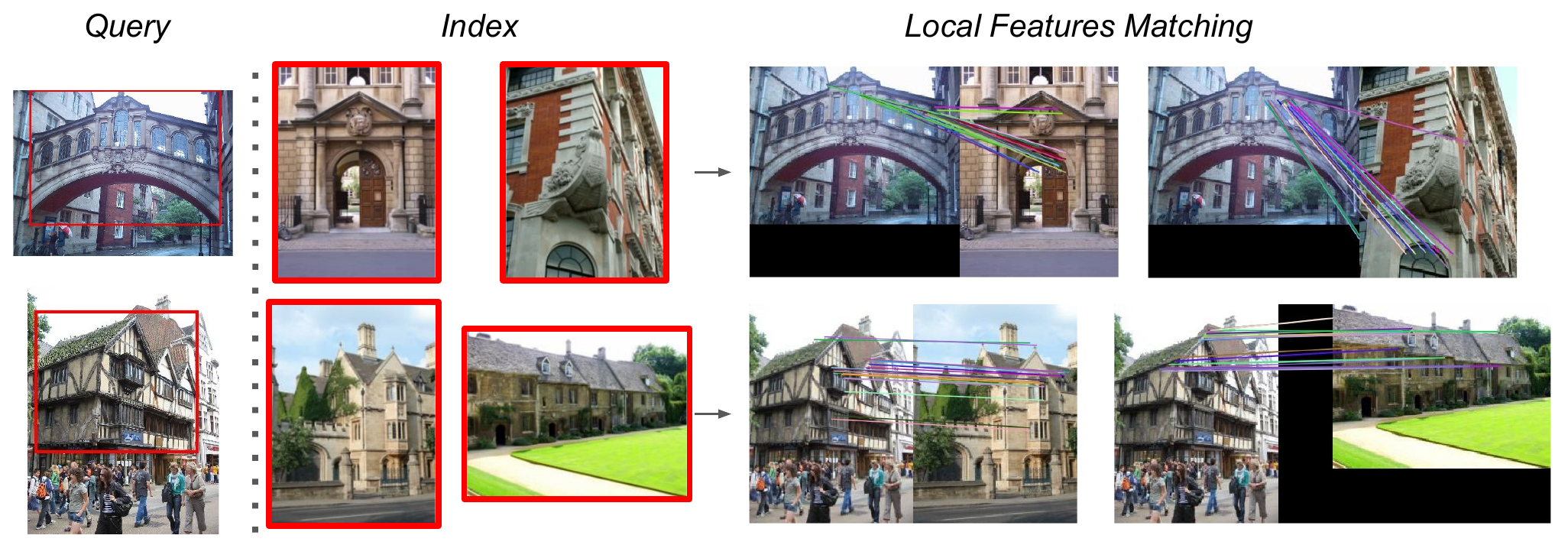}
\caption{Examples of incorrect local feature matches, for image pairs depicting different objects/scenes.}
\label{fig:negative_matching}
\end{figure*}

We present more qualitative results, to illustrate local feature matching with DELG: \figref{fig:positive_matching} presents examples of correct matches, and \figref{fig:negative_matching} presents examples of incorrect matches.
For these examples, we use the R50-DELG model and the \roxf{}/\rpar{} datasets.
These visualizations depict the final obtained correspondences, post-RANSAC.

For each row, one query and two index images are shown, and on the right their local feature matches are shown, with lines connecting the corresponding keypoints.
\figref{fig:positive_matching} showcases DELG's robustness against strong viewpoint and illumination changes: for example, matches can be obtained across different scales and day-vs-night cases.
\figref{fig:negative_matching} presents overtriggering cases, where a match is found even though different objects/scenes are presented: these tend to occur for similar patterns between query and index images (ie, similar windows, arches or roofs) which appear in similar geometric configurations.
Generally, these do not affect retrieval results much because the number of inliers is low.

%% file: sections/viz_dataset_examples.tex

\section*{Appendix F. Feature visualization}

We provide visualizations of the features learned by the DELG model.
This is useful to understand the hierarchical representation which we rely on for extraction of local and global features.
We explore two types of visualizations, based on dataset examples with high activations and by optimizing input images with respect to a given layer/channel.
 
\subsection*{F.1 Feature visualization by dataset examples}

For this experiment, we run our R50-DELG model over $200$k images from the Google Landmarks dataset \cite{noh2017large}, and collect the images and feature positions with largest value in each channel of several activation maps.
We specifically consider the activation maps at the outputs of the \textit{conv2}, \textit{conv3}, \textit{conv4} and \textit{conv5} blocks of layers in our model's ResNet architecture \cite{he2015deep}.
The feature positions with maximum activations can be mapped back to the relevant input image regions by computing the model's receptive field parameters \cite{araujo2019computing}.
Note that the region may partly fall outside the image, in which case we apply zero-padding for the visualization.

\figref{fig:dataset_examples} presents image patches that produce the highest activations for selected channels of the above-mentioned layers.
The activation values are noted in each subfigure's title and can be read by zooming in.
For each selected channel of a specific layer, the $9$ patches with largest activations are shown.
The receptive field sizes (both horizontally and vertically) for each of these layers are: $35$ (\textit{conv2}), $99$ (\textit{conv3}), $291$ (\textit{conv4}) and $483$ (\textit{conv5}); these correspond to the sizes of image patches.
One can notice that the types of patterns which maximally activate specific layers grow in complexity with network depth.
This agrees with observations from previous work, where the hierarchical nature of CNN features is discussed \cite{zeiler2014visualizing,olah2017feature}.
Shallower layers such as \textit{conv2} tend to focus on edges and simple textures; \textit{conv3} responds highly to more complex shapes, such as edges resembling palm trees (channel $15$) and arches (channel $48$); \textit{conv4} focuses on object parts, such as green dome-like shapes (channel $2$), or arches (channel $30$); \textit{conv5} shows strong activations for entire objects, with substantial invariances to viewpoint and lighting: entire buildings (channel $3$), islands (channel $23$) or towers (channel $52$) are captured.

These visualizations help with intuitive understanding of our proposed method, which composes local features from a shallower layer (\textit{conv4}) and global features from a deeper layer (\textit{conv5}).
Features from \textit{conv5} present high degree of viewpoint invariance, being suboptimal for localized matching and more suitable to global representations.
In contrast, \textit{conv4} features seem more grounded to localizable object parts and thus can be effectively used as local feature representations.

%% file: sections/viz_optimization.tex

\subsection*{F.2 Feature visualization by optimization}

In this experiment, we consider the same layers and channels as above, but now adopt a visualization technique by optimizing the input image to maximally activate the desired feature. 
First, the input image is initialized with random noise.
Given the desired layer/channel, we backpropagate gradients in order to maximally activate it.
Regularizers can be useful to restrict the optimization space, otherwise the network may find ways to activate neurons that don't occur in natural images.

\begin{figure*}
\centering
\begin{tabular}{ccc}

\includegraphics[height=0.2 \textwidth]{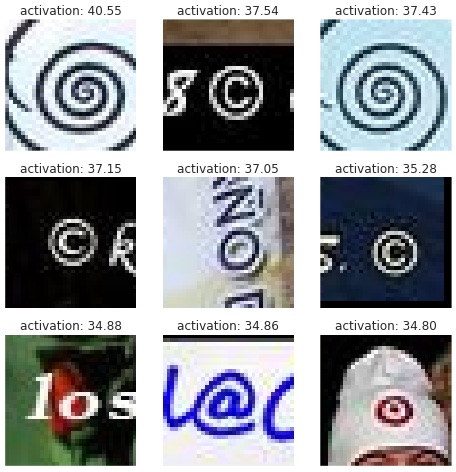}&
\includegraphics[height=0.2 \textwidth]{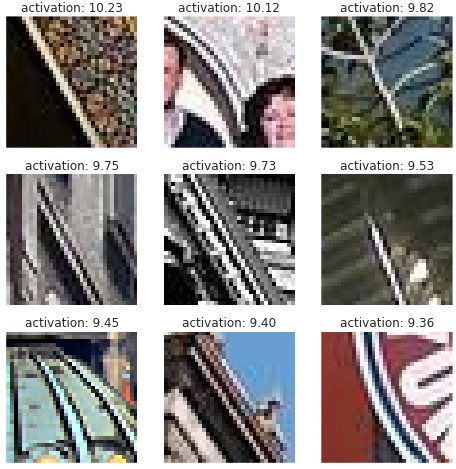}&
\includegraphics[height=0.2 \textwidth]{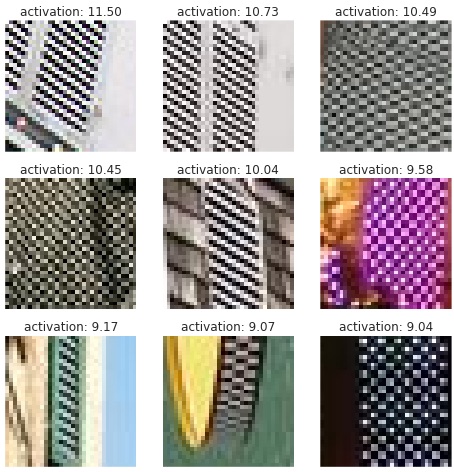} \\
\textit{conv2}, channel $0$ & \textit{conv2}, channel $12$ & \textit{conv2}, channel $49$\\
\\
\includegraphics[height=0.25 \textwidth]{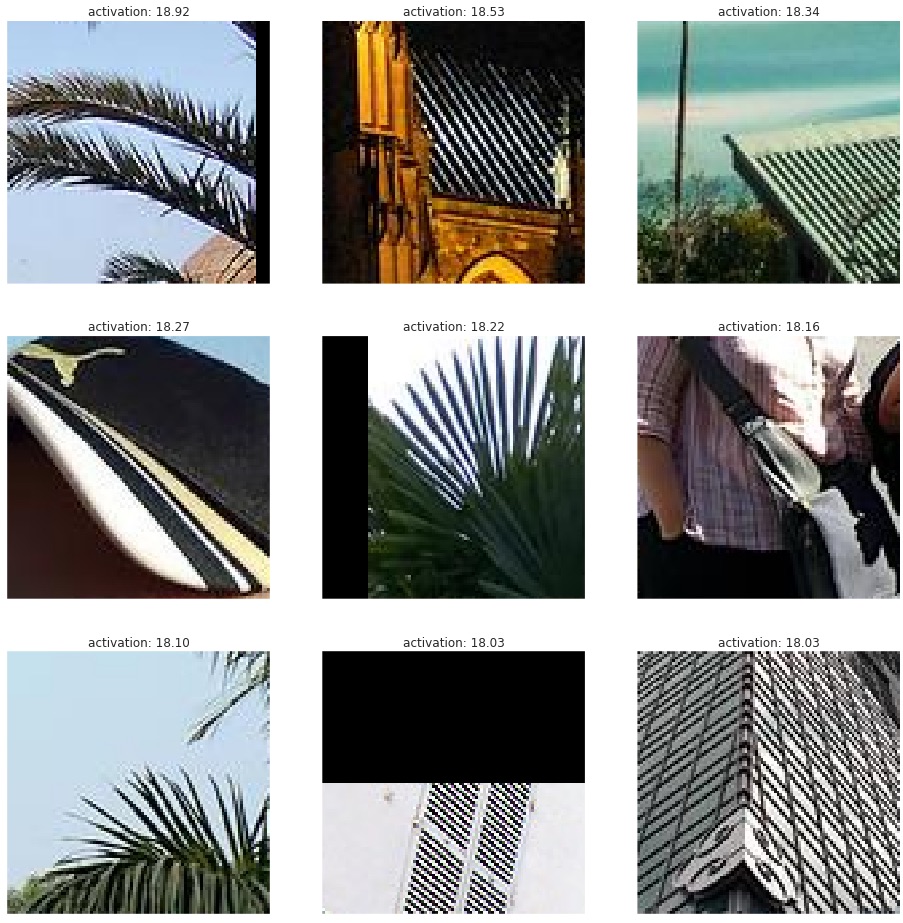}&
\includegraphics[height=0.25 \textwidth]{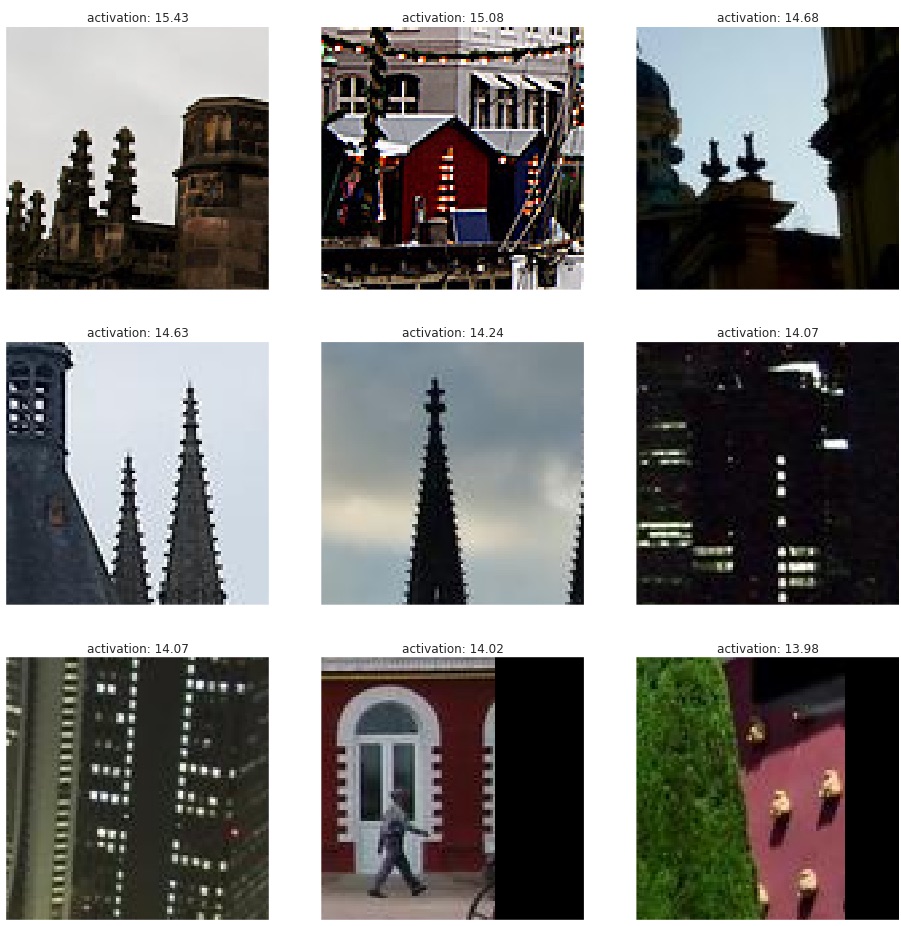}&
\includegraphics[height=0.25 \textwidth]{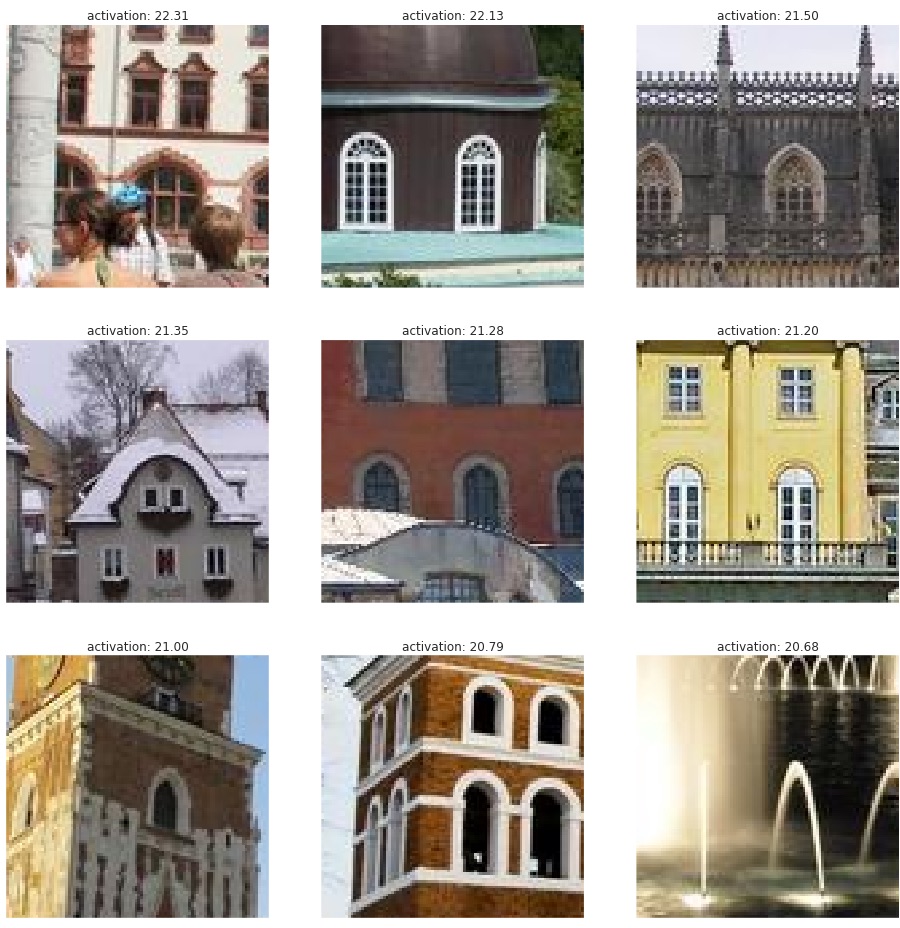} \\
\textit{conv3}, channel $15$ & \textit{conv3}, channel $29$ & \textit{conv3}, channel $48$ \\
\\
\includegraphics[height=0.3 \textwidth]{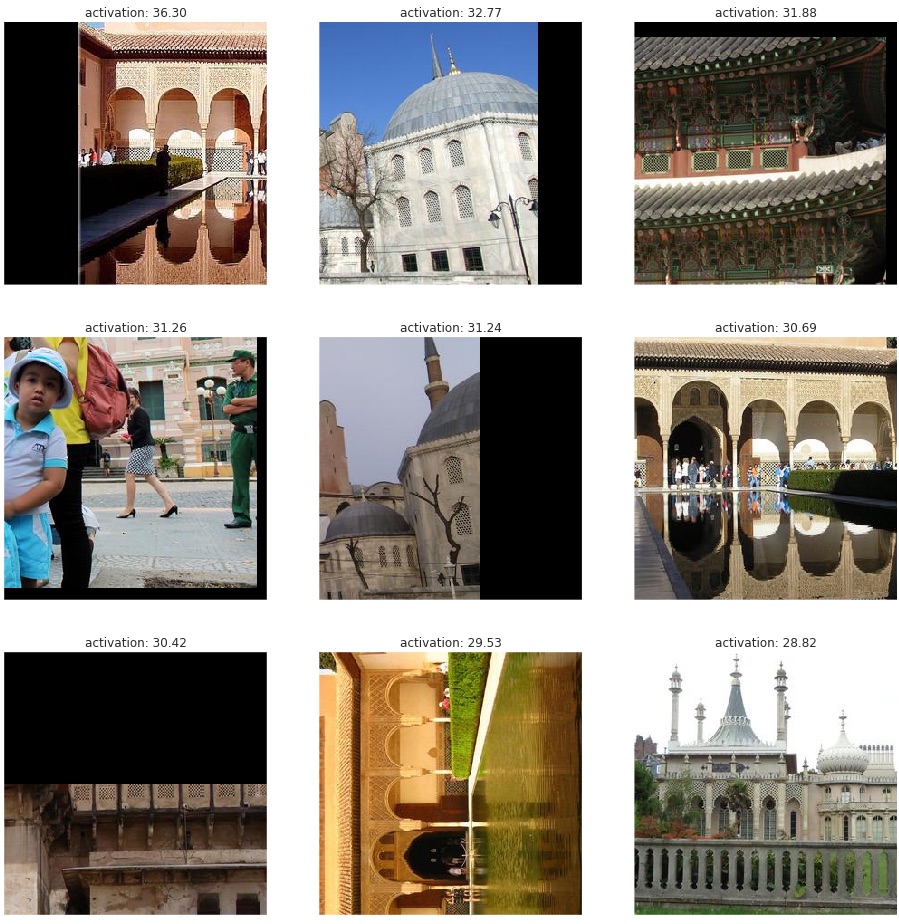}&
\includegraphics[height=0.3 \textwidth]{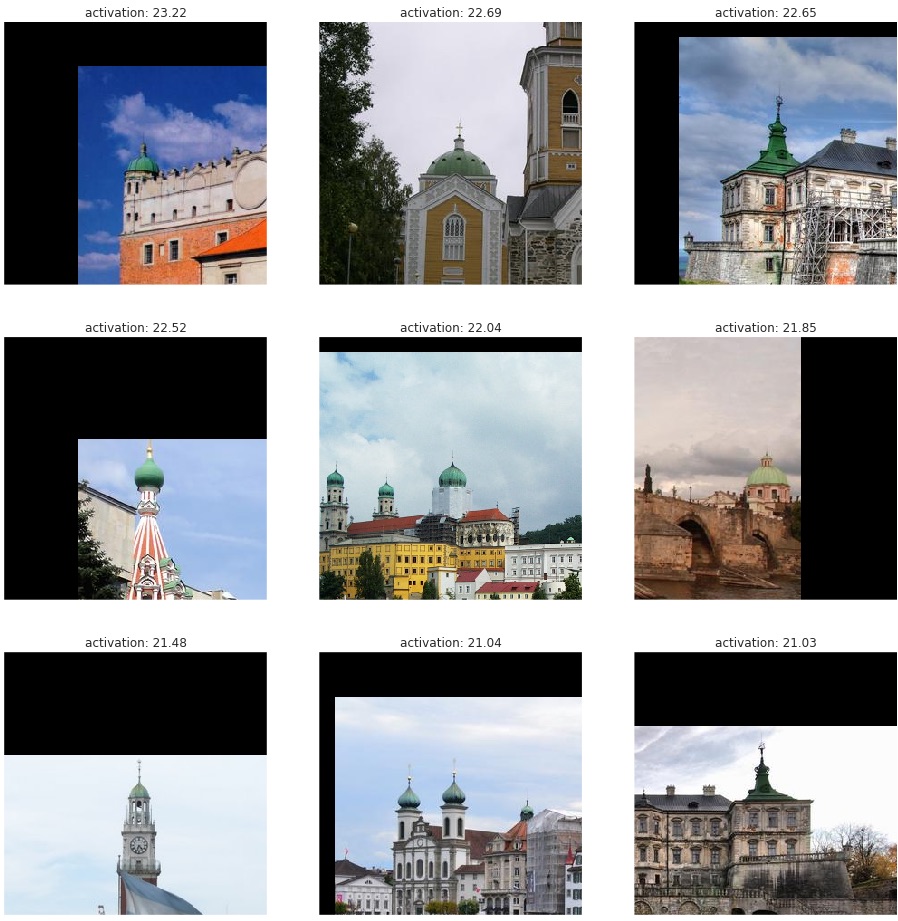}&
\includegraphics[height=0.3 \textwidth]{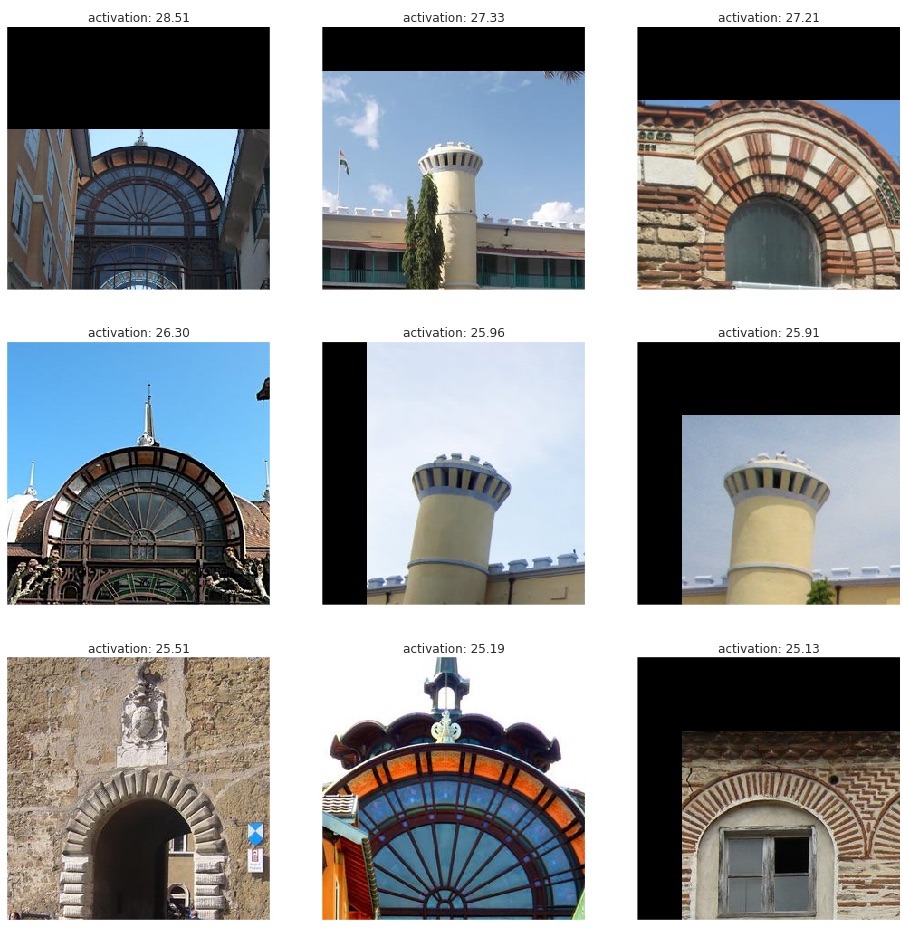} \\
\textit{conv4}, channel $1$ & \textit{conv4}, channel $2$ & \textit{conv4}, channel $30$ \\
\\
\includegraphics[height=0.3 \textwidth]{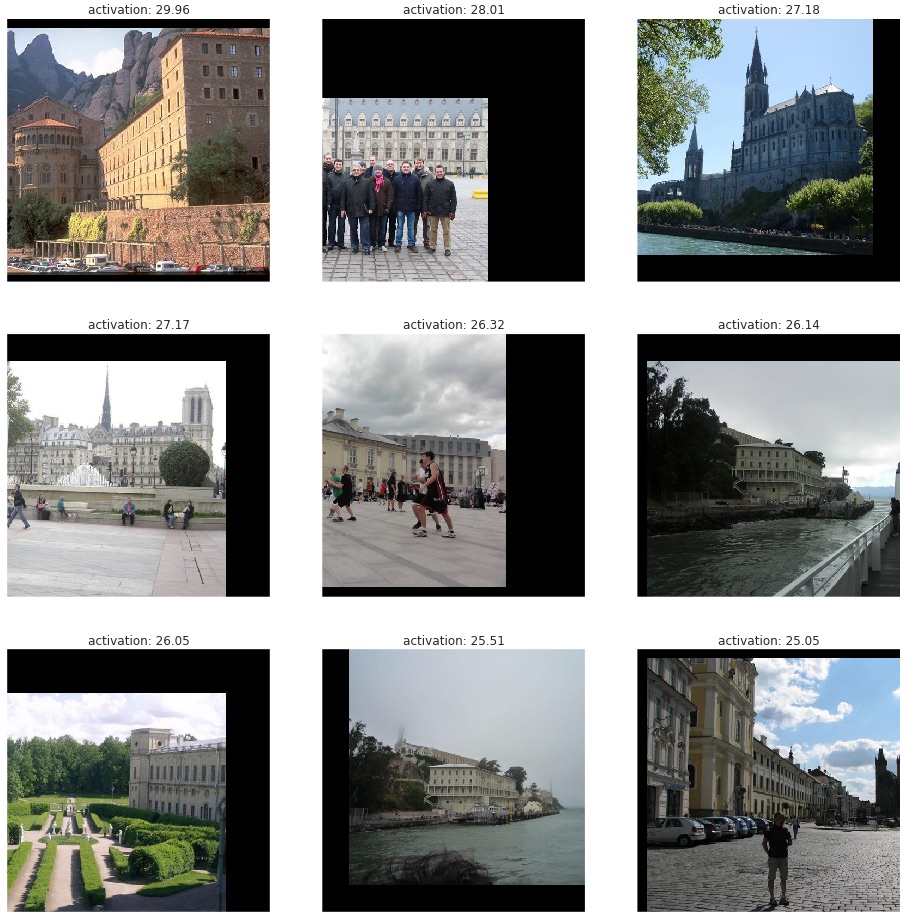}&
\includegraphics[height=0.3 \textwidth]{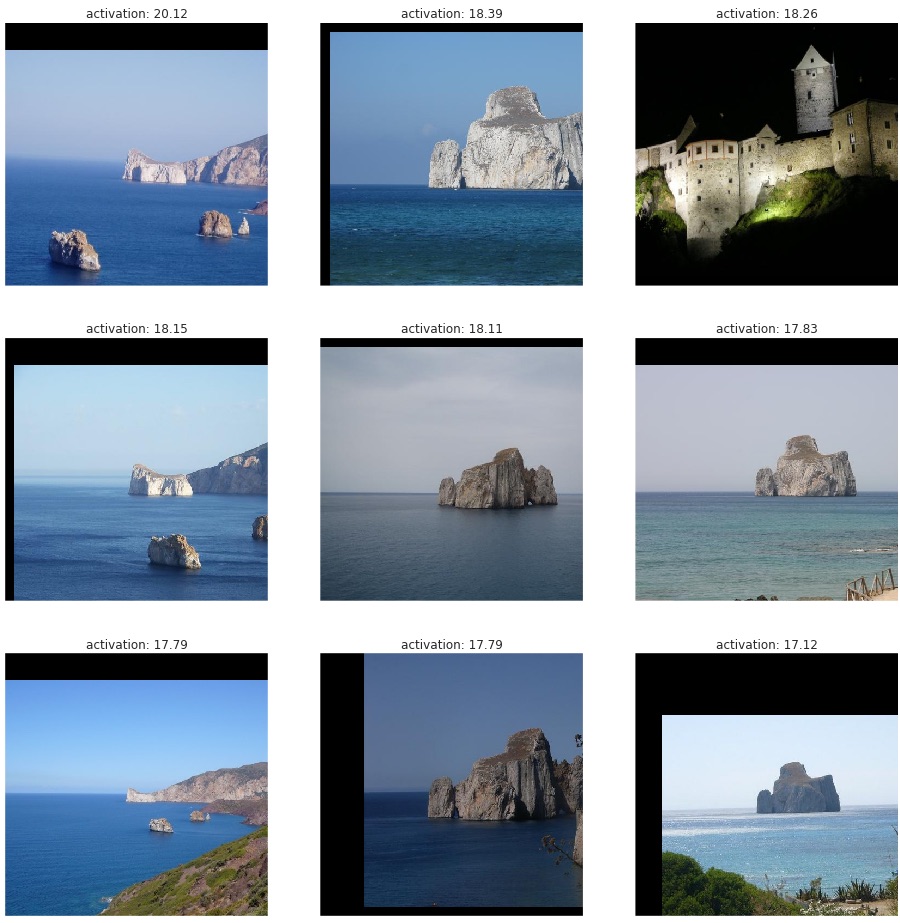}&
\includegraphics[height=0.3 \textwidth]{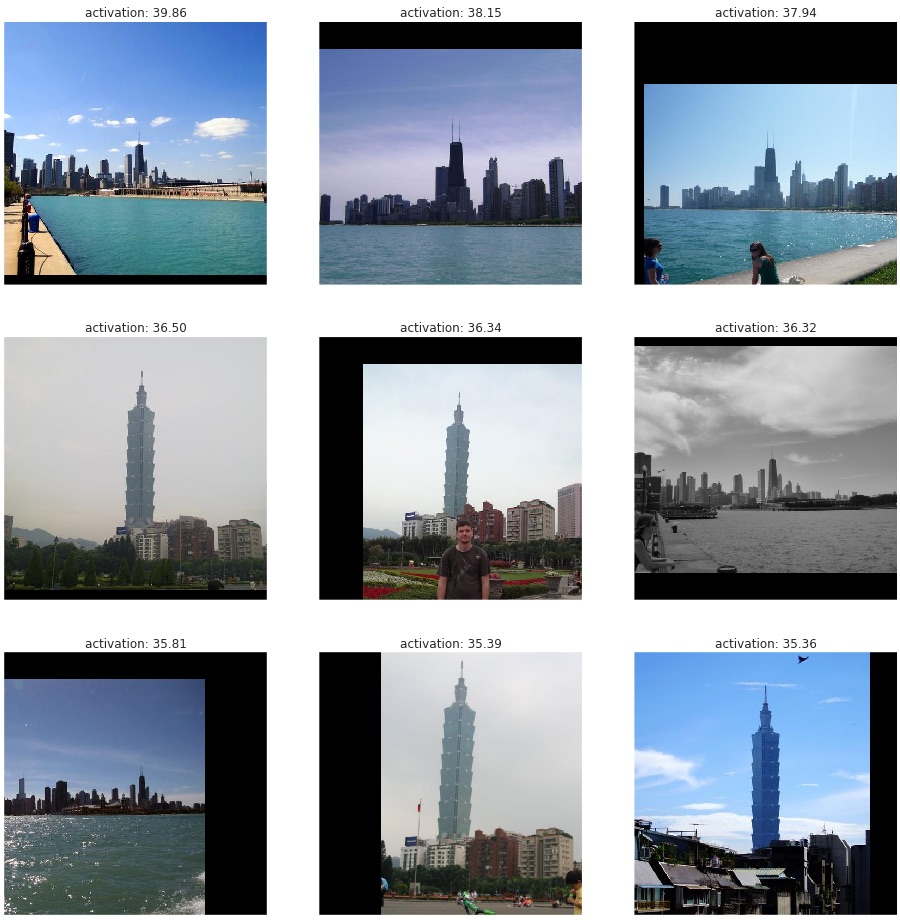} \\
\textit{conv5}, channel $3$ & \textit{conv5}, channel $23$ & \textit{conv5}, channel $52$ \\
\end{tabular}
\caption{Visualization of patterns detected by specific feature maps / channels, by presenting image patches that produce high activations.}
\label{fig:dataset_examples}
\end{figure*}

\begin{figure*}
\centering
\begin{tabular}{ccc}

\includegraphics[height=0.25 \textwidth]{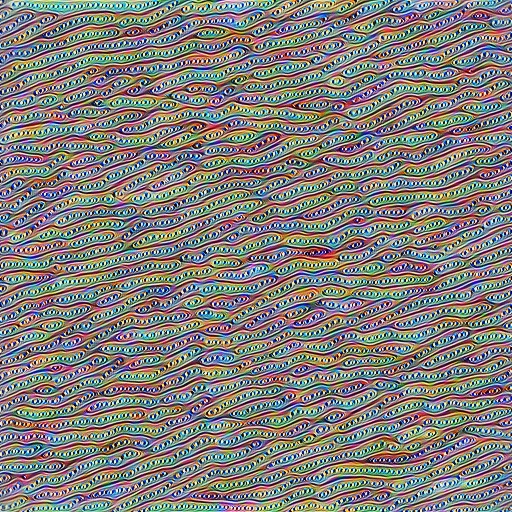}&
\includegraphics[height=0.25 \textwidth]{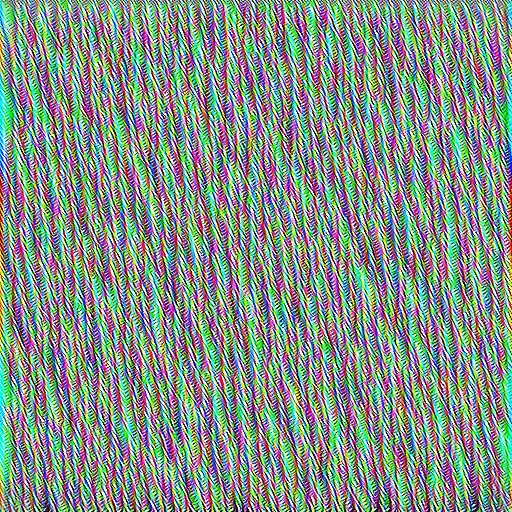}&
\includegraphics[height=0.25 \textwidth]{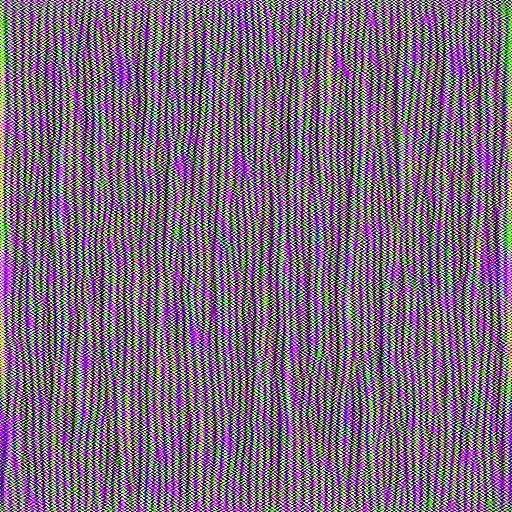} \\
\textit{conv2}, channel $0$ & \textit{conv2}, channel $12$ & \textit{conv2}, channel $49$\\
\\
\includegraphics[height=0.25 \textwidth]{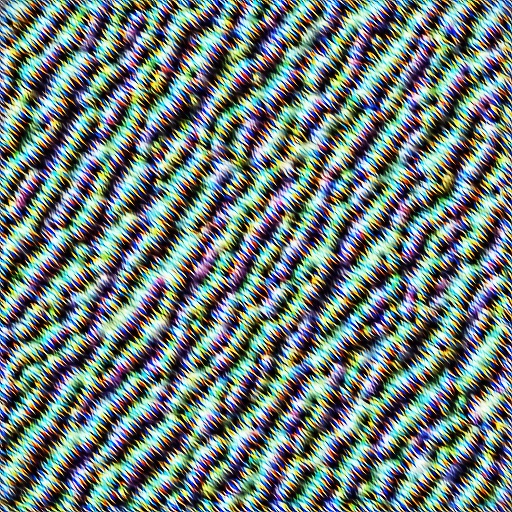}&
\includegraphics[height=0.25 \textwidth]{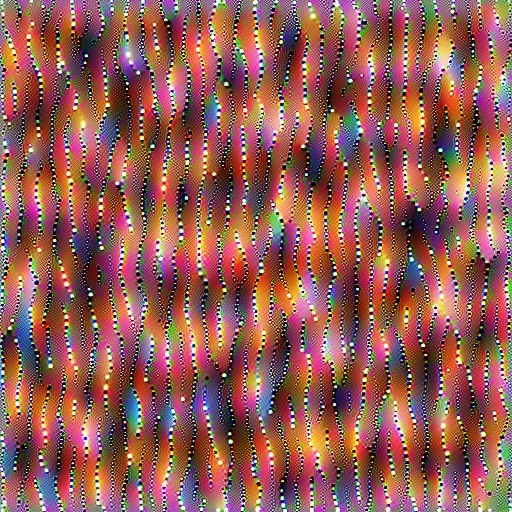}&
\includegraphics[height=0.25 \textwidth]{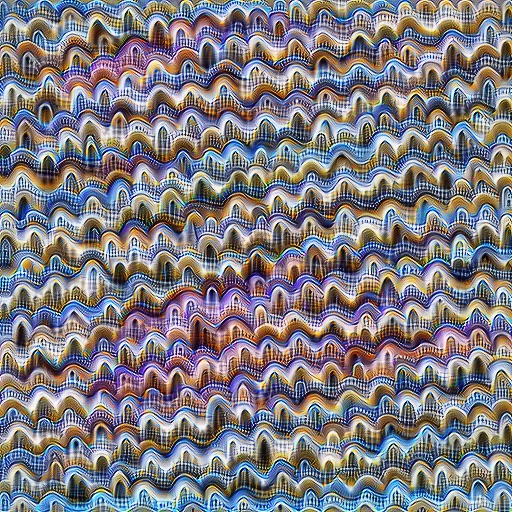} \\
\textit{conv3}, channel $15$ & \textit{conv3}, channel $29$ & \textit{conv3}, channel $48$ \\
\\
\includegraphics[height=0.25 \textwidth]{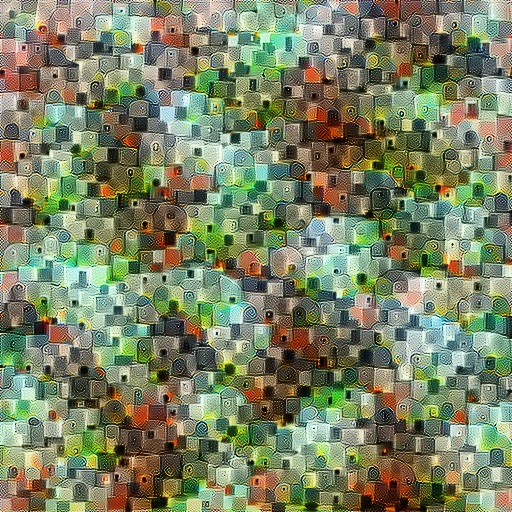}&
\includegraphics[height=0.25 \textwidth]{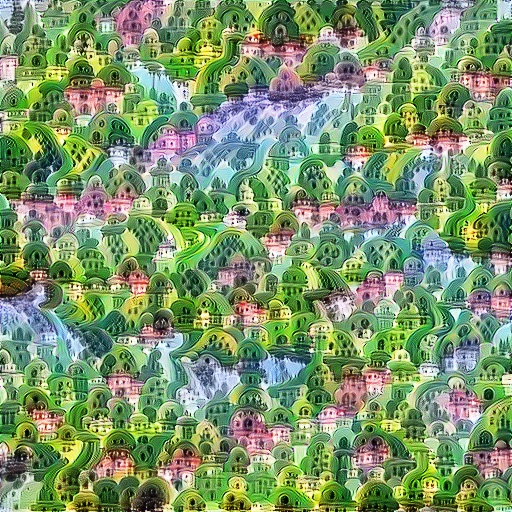}&
\includegraphics[height=0.25 \textwidth]{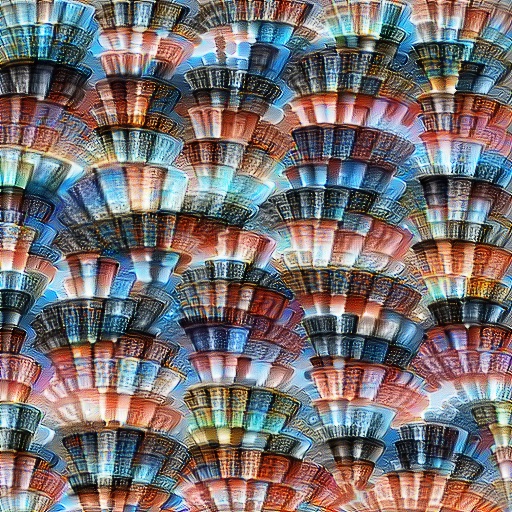} \\
\textit{conv4}, channel $1$ & \textit{conv4}, channel $2$ & \textit{conv4}, channel $30$ \\
\\
\includegraphics[height=0.25 \textwidth]{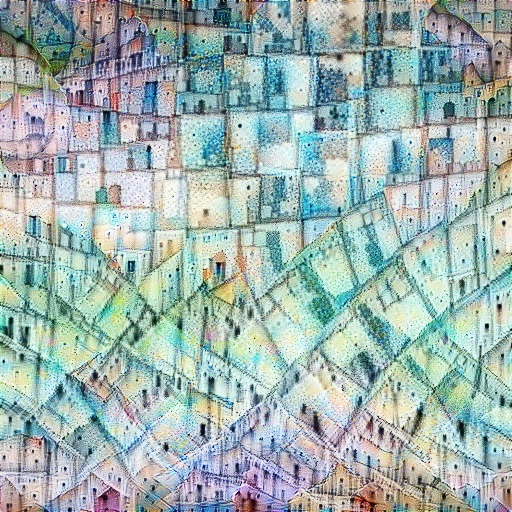}&
\includegraphics[height=0.25 \textwidth]{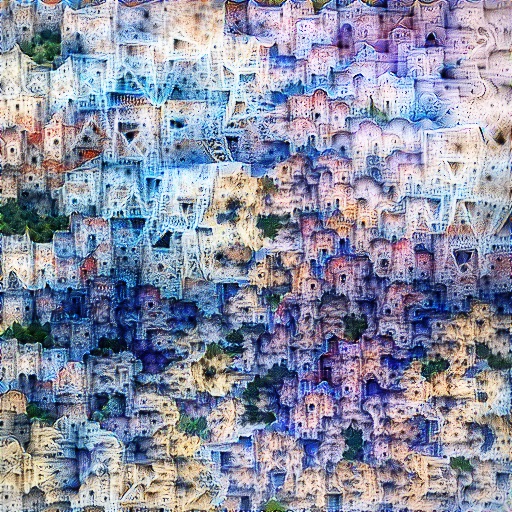}&
\includegraphics[height=0.25 \textwidth]{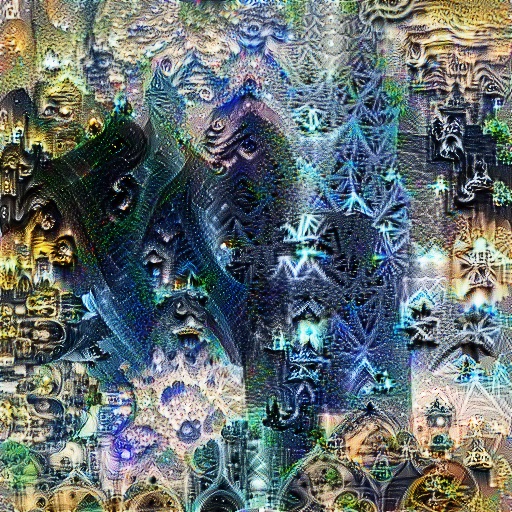} \\
\textit{conv5}, channel $3$ & \textit{conv5}, channel $23$ & \textit{conv5}, channel $52$ \\
\end{tabular}
\caption{Visualization of patterns detected by specific feature maps / channels, by optimizing the input image to produce high activations.}
\label{fig:optimization}
\end{figure*}

We reuse the technique from Olah \etal \cite{olah2017feature}, with default parameters, and the results are presented in \figref{fig:optimization}.
Again, we notice that a hierarchical representation structure forms, with more complex patterns being produced as the network goes deeper from \textit{conv2} to \textit{conv5}.
As expected, the produced images agree very well with the patches from \figref{fig:dataset_examples}, in terms of the types of visual contents.
For example, for \textit{conv3} channel $48$, the optimized image shows arch-like edges while the dataset examples present image patches where those types of patterns occur.

Note also how deeper layers tend to specialize for the target task, by detecting image patches with parts and texture that are common to landmarks.
For example, \textit{conv4} shows detection of green dome-like shapes (channel $2$) and arches (channel $30$), which are common in these types of objects; \textit{conv5}, on the other hand, shows building walls (channel $3$) and rocky patterns that are common in ancient buildings or islands (channel $23$). 


